\documentclass{article} % For LaTeX2e
\usepackage{iclr2026_malgai,times}

\usepackage{amsmath,amsfonts,bm}

\def\eqref#1{equation~\ref{#1}}
\def\1{\bm{1}}

\DeclareMathAlphabet{\mathsfit}{\encodingdefault}{\sfdefault}{m}{sl}
\SetMathAlphabet{\mathsfit}{bold}{\encodingdefault}{\sfdefault}{bx}{n}

\usepackage[backref=page]{hyperref}
\usepackage{url}
\usepackage{booktabs}
\usepackage{multirow}
\usepackage{graphicx}
\usepackage[table]{xcolor}
\usepackage{subcaption}
\usepackage{float}
\usepackage{pifont}

\newcommand{\cmark}{\ding{51}}
\newcommand{\xmark}{\ding{55}}

\title{Cattle Trade: A Multi-Agent Benchmark for LLM Bluffing, Bidding, and Bargaining}

\author{Robert M\"uller\thanks{Equal contribution. Correspondence to: \texttt{robert@aganthos.com}}
\And
Clemens M\"uller\footnotemark[1]}

\iclrfinalcopy  % camera-ready: reveals authors, removes line numbers
\begin{document}

\maketitle

\begin{abstract}
We introduce \textsc{Cattle Trade}\footnote{\emph{Kuhhandel} (English: \emph{You're Bluffing!}) is a
trademark of Ravensburger. This work is not affiliated with Ravensburger.}, a multi-agent benchmark
for evaluating large language models (LLMs) as agents in strategic reasoning under imperfect
information, adversarial interaction, and resource constraints. The benchmark combines auctions,
hidden-offer trade challenges (TCs), bargaining, bluffing, opponent modeling, and
resource allocation within a single long-horizon game lasting 50--60 turns. Unlike
prior agent benchmarks that test these abilities in isolation, \textsc{Cattle Trade} evaluates whether
agents integrate them across a competitive, multi-agent economic game with conflicting incentives.
The benchmark logs every bid, TC offer, counteroffer, and card selection, enabling behavioural
analysis beyond final scores or win rates. We evaluate seven cost-efficient language models and
three deterministic code agents across 242 games. Strategic coherence, in particular spending efficiency, resource
discipline, and phase-adaptive bidding, is associated with rank more strongly than spending volume or any single
subskill. Two heuristic code agents outperform most tested LLMs, and behavioural traces surface
recurring LLM failure modes including overbidding, self-bidding, bankrupt TC initiation, and
weak opponent-state adaptation. Evaluating agentic competence requires benchmarks that test the
joint deployment of multiple capabilities in multi-agent environments with conflicting incentives,
uncertainty, and economic dynamics.
\end{abstract}

% Introduction
\section{Introduction}

LLM-based agents are increasingly deployed in multi-agent settings, from
negotiation~\citep{lewis2017deal} to cooperative tool use~\citep{hong2024metagpt}. Standard
benchmarks such as MMLU~\citep{hendrycks2021measuring} and HumanEval~\citep{chen2021evaluating}
measure knowledge and code generation but tell us little about strategic reasoning under uncertainty
and adversarial incentives. Game-based benchmarks for social
deduction~\citep{light2023avalonbench,xu2023werewolf}, game-theoretic
scenarios~\citep{duan2024gtbench}, and social reasoning~\citep{pan2023machiavelli} each test
individual competencies in isolation rather than requiring their combination within a single
environment.

% --- Figure 1 (decision-tree + ranking table) 
\begin{figure}[h]
\centering
\begin{minipage}[c]{0.68\linewidth}
\centering
\includegraphics[width=\linewidth]{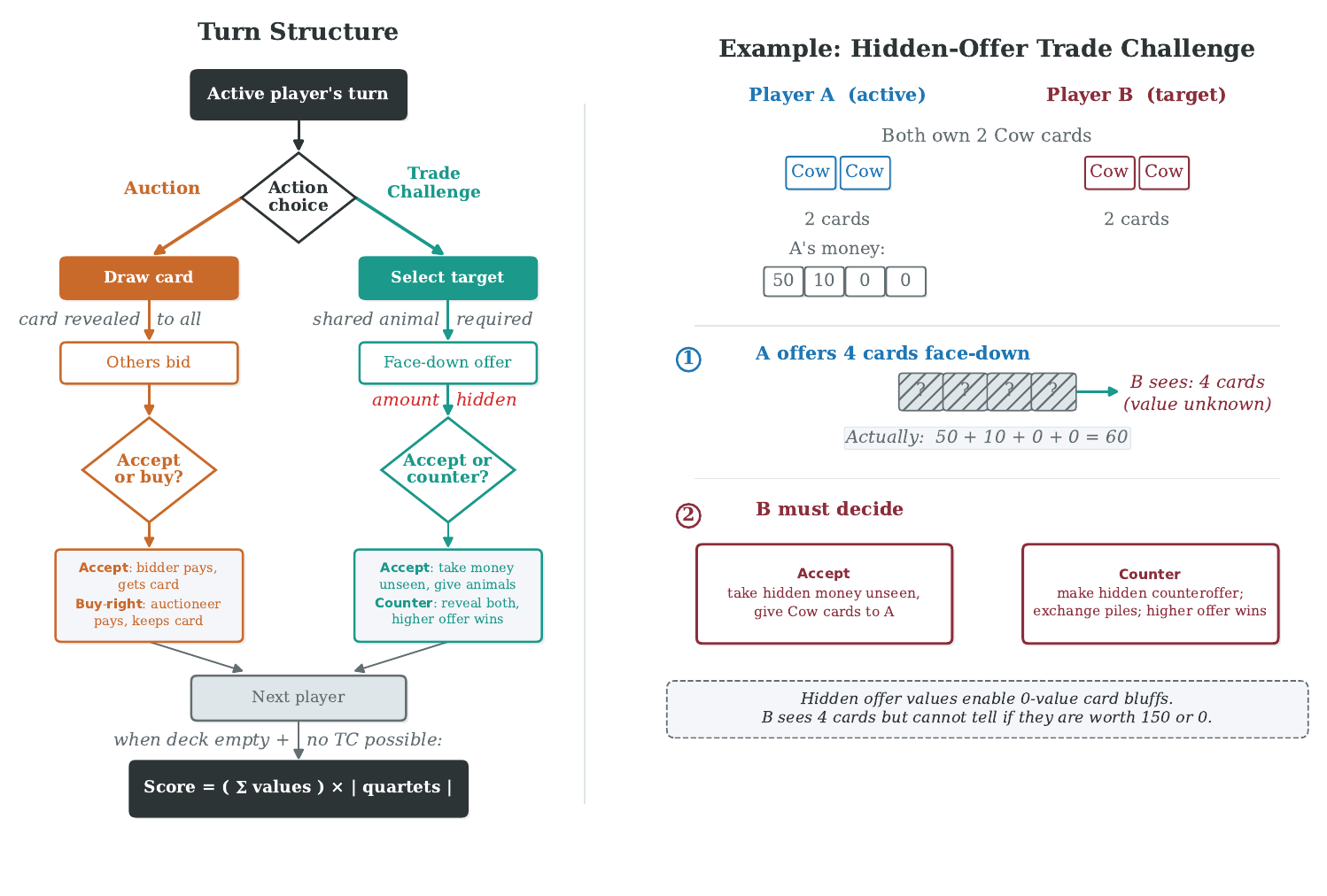}
\end{minipage}\hfill
\begin{minipage}[c]{0.3\linewidth}
\centering
\scriptsize
\setlength{\tabcolsep}{2pt}
\renewcommand{\arraystretch}{1.05}
\definecolor{llmrow}{HTML}{FFF4CC}
\definecolor{coderow}{HTML}{D6E9F8}
\begin{tabular}{@{}lcc@{}}
\toprule
Agent & TS $\mu\!\pm\!\sigma$ & Median\,$\pm$\,std \\
\midrule
\rowcolor{llmrow}  G3-F            & $30.1\!\pm\!1.1$ & $5{,}250\!\pm\!4{,}026$ \\
\rowcolor{coderow} C:Trk           & $28.7\!\pm\!1.2$ & $3{,}550\!\pm\!2{,}688$ \\
\rowcolor{llmrow}  G3.1-FL         & $28.0\!\pm\!1.0$ & $3{,}930\!\pm\!2{,}652$ \\
\rowcolor{coderow} C:Set           & $27.3\!\pm\!1.1$ & $2{,}760\!\pm\!1{,}555$ \\
\rowcolor{llmrow}  Sonnet 4.5  & $26.4\!\pm\!1.6$ & $2{,}310\!\pm\!2{,}376$ \\
\rowcolor{llmrow}  DS-v3.2         & $23.9\!\pm\!0.9$ & $1{,}640\!\pm\!1{,}858$ \\
\rowcolor{llmrow}  GPT5.4-N        & $22.6\!\pm\!0.9$ & $1{,}490\!\pm\!2{,}667$ \\
\rowcolor{coderow} C:Econ          & $22.1\!\pm\!1.2$ & $1{,}000\!\pm\!1{,}110$ \\
\rowcolor{llmrow}  Haiku 4.5          & $21.8\!\pm\!0.9$ & $1{,}000\!\pm\!1{,}596$ \\
\rowcolor{llmrow}  G2.5-FL         & $20.5\!\pm\!1.0$ & $\phantom{0}500\!\pm\!1{,}283$ \\
\bottomrule
\end{tabular}\\[2pt]
{\tiny\textcolor{gray}{\colorbox{llmrow}{\strut LLM} \colorbox{coderow}{\strut Code}}}
\end{minipage}
\vspace{-6pt}
\caption{\textbf{(a) Turn structure and hidden-offer trade challenge.} Card counts are public; offer values
are hidden, enabling 0-value card bluffs. \textbf{(b) Ranking} over 98 canonical games sorted by
TrueSkill $\mu$; median $\pm$ std in the right column. TrackerAgent and SetRaceAgent beat
six and five of seven LLMs, respectively.}
\label{fig:gameflow}
\end{figure}

Passing individual tests does not guarantee joint competence. One key challenge in evaluating agents is not whether models possess isolated skills, but whether they can integrate them coherently under uncertainty, competition, and resource allocation constraints. This matters for downstream agentic settings such as negotiation, procurement, portfolio management and other economic interactions, where effective behavior depends on combining tasks like bidding, bargaining, bluffing, and resource management over time. Prior benchmarks often probe subsets of these abilities, but rarely test whether agents can deploy them jointly in a strategically structured adversarial environment.

We introduce \textsc{Cattle Trade}, a benchmark that integrates
(i)~competitive auctions with a buy-right mechanism coupling bidding strategy to future liquidity,
(ii)~hidden-offer bilateral trade challenges (TCs) that create bluffing incentives, (iii)~discrete money with
no-change payments that impose resource constraints, and (iv)~a multiplicative scoring rule that
makes portfolio composition consequential. Games last 50--60 turns with four players. We evaluate
seven cost-efficient LLMs and three deterministic code agents across 242 games in two formats:
pure-LLM tournaments and mixed games where each LLM faces three code agents.

TrackerAgent and SetRaceAgent outperform six and five of seven LLMs respectively. Only Gemini~3 Flash clearly
outperforms all code-agent baselines (72.9\% wins, $\mu=30.1\!\pm\!3.3$); TrackerAgent
($\mu=28.7\!\pm\!3.6$) and Gemini~3.1 Flash Lite ($\mu=28.0\!\pm\!2.9$) sit close behind,
with G3.1-FL clearing SetRace and EconomyAgent but essentially tied with TrackerAgent.
Behavioural analysis surfaces failures absent from the code agents: overbidding, self-bidding
spirals, and bankrupt TC initiation.

Our contributions:
\begin{itemize}
\item A multi-agent benchmark combining auctions, bargaining, deception, and resource constraints
    with structured decision logging and configurable game parameters.
\item A behavioural analysis suite profiling strategic play via spending efficiency, bluff rates,
    phase-dependent bid adaptation, self-bidding rates, and buy-right patterns.
\item Empirical results across 242 games showing that spending efficiency and resource discipline
    is associated with success, while overbidding, self-bidding spirals, and undisciplined bluffing characterise
    failure.
\end{itemize}

% Related work
\section{Related Work}
\label{sec:related}

Standard benchmarks such as MMLU~\citep{hendrycks2021measuring},
BIG-bench~\citep{srivastava2023beyond}, and HumanEval~\citep{chen2021evaluating} measure knowledge
and reasoning but do not test multi-agent incentives, hidden information, or long-horizon resource
allocation. AgentBench~\citep{liu2023agentbench} extends evaluation to interactive tool-use settings
but lacks adversarial strategic reasoning.

Game-based LLM benchmarks each emphasize a subset of strategic skills.
GTBench~\citep{duan2024gtbench} covers game-theoretic scenarios but uses short-horizon interactions.
AvalonBench~\citep{light2023avalonbench} and Werewolf~\citep{xu2023werewolf} probe deception and
cooperation but lack economic constraints. MACHIAVELLI~\citep{pan2023machiavelli} provides
long-horizon continuous scoring but pits a single agent against scripted dynamics.
Suspicion-Agent~\citep{guo2024suspicion} studies deception in an imperfect-information board game;
GameBench~\citep{costarelli2024gamebench} broadens game coverage but simplifies economic structure;
lmgame-Bench~\citep{hu2026lmgamebench} proposes a Gym-style interface with scaffolds for perception
and memory. To our knowledge, no existing benchmark requires agents to combine auctions,
hidden-offer trade challenges, bluffing, and discrete resource management within a single game.

\citet{jia2025llmstrategic} analyze strategic reasoning through behavioral game theory,
GAMEBoT~\citep{lin2025gamebot} decomposes game reasoning into interpretable sub-problems, and
ALYMPICS~\citep{mao2025alympics} provides an empirical game-theory framework whose pilot task
centres on scarce resource allocation through multi-round auctions.
Pluribus~\citep{brown2019superhuman} and CICERO~\citep{bakhtin2022human} achieve strong play in
poker and Diplomacy, respectively, but rely on domain-specific search and self-play rather than
general-purpose language modeling. Work on neural
negotiation~\citep{lewis2017deal,gandhi2023strategic} and deception~\citep{hagendorff2024deception}
addresses dialogue-level strategy. 

These works are closest in
their concern with strategic interaction, but they do not test the joint deployment of auctions,
hidden-offer trade challenges, bluffing, discrete no-change payments, and portfolio-sensitive scoring.

\textsc{Cattle Trade} concentrates these pressures into a single environment: a multi-agent
imperfect-information game with repeated auctions and bargaining, bluffing incentives, discrete
liquidity constraints, and continuous scoring over 50--60 turns. This enables both head-to-head
ranking (via TrueSkill) and mechanism-level behavioural diagnosis (overbid frequency, self-bidding
rate, bluff calibration).

% Cattle Trade — game rules, benchmark design, agent interface
\section{Cattle Trade}

\subsection{The Cattle Trade Game}

\textsc{Cattle Trade} adapts a 3--5 player tabletop card game designed by R\"udiger
Koltze (Ravensburger, 1985). Players collect quartets (complete sets of 4 cards) of animal types,
scored by a multiplicative formula that rewards both the value and number of completed sets. The
game uses 40 animal cards (10 types $\times$ 4 each) and 55 money cards of fixed denominations.
Appendix~\ref{app:rules} details the turn structure and TC mechanic;
Table~\ref{tab:comparison} compares \textsc{Cattle Trade} to existing game-based LLM benchmarks.
\textsc{Cattle Trade} combines auctions, hidden-offer trade challenges (TCs), bluffing, and
resource management within a single game over
50--60 turns.

\begin{table}[h]
\centering
\setlength{\tabcolsep}{4pt}
\resizebox{\textwidth}{!}{%
\begin{tabular}{lcccccccccc}
\toprule
Benchmark & Players & Info & Bargain & Bluff & Econ. & Hidden & Discrete\$ & Buy-right & Scoring & Turns \\
\midrule
GTBench & 2 & Mixed & \cmark & \xmark & \xmark & \xmark & \xmark & \xmark & Binary & 1--10 \\
AvalonBench & 5--10 & Imperfect & \xmark & \cmark & \xmark & \xmark & \xmark & \xmark & Binary & 5--10 \\
Werewolf & 5--18 & Imperfect & \xmark & \cmark & \xmark & \xmark & \xmark & \xmark & Binary & 5--15 \\
MACHIAVELLI & 1 & Partial & \xmark & \xmark & \xmark & \xmark & \xmark & \xmark & Continuous & 50+ \\
Suspicion-Agent & 3--6 & Imperfect & \xmark & \cmark & \xmark & \xmark & \xmark & \xmark & Binary & 10--20 \\
ALYMPICS & 5 & Imperfect & \xmark & \xmark & \cmark & \xmark & \xmark & \xmark & Continuous & 5--10 \\
\midrule
\textsc{Cattle Trade} & 3--5 & Imperfect & \cmark & \cmark & \cmark & \cmark & \cmark & \cmark & Continuous & 50--60 \\
\bottomrule
\end{tabular}%
}
\caption{Comparison of game-based LLM benchmarks. Bargain: explicit bilateral offer/counter mechanic. Bluff: deception as core mechanic. Econ.: resource management required. Hidden: face-down offers create information asymmetry within bargaining (an opponent must accept or counter without seeing the offered amount). Discrete\$: payments use fixed-denomination money cards with no change given, so paying 60 with a 100 card costs 100. Buy-right: the auctioneer may pay the highest bid to keep the auctioned card rather than sell. Suite-style benchmarks (GameBench, GAMEBoT, lmgame-Bench) cover broad ranges of games whose properties vary by game; we cite them in \S\ref{sec:related} but omit from this table because per-property values would not be representative. \textsc{Cattle Trade} combines these properties through the TC mechanic and auction phase.}
\label{tab:comparison}
\end{table}

Each animal type has a quartet value: Chicken (10), Goose (40), Cat (90), Dog (160), Sheep (250),
Goat (350), Donkey (500), Pig (650), Cow (800), and Horse (1000). The final game score is:
\begin{equation}
    \text{Score} = \left(\sum_{q \in \mathcal{Q}} v_q\right) \times |\mathcal{Q}|
    \label{eq:scoring}
\end{equation}
where $\mathcal{Q}$ is the set of completed quartets and $v_q$ is the value of quartet $q$. This
multiplicative structure means three mid-value quartets (e.g., 1{,}250 $\times$ 3 = 3{,}750)
outscore two top quartets (1{,}800 $\times$ 2 = 3{,}600), rewarding breadth over concentration and
making opponent modeling critical.

The game uses discrete, non-fungible money cards (denominations 0, 10, 50, 100, 200, 500) with no
change given: paying 60 with a 100 card costs 100. Each player starts with 90 coins (four 10s, one
50, two 0s). The 0 cards exist specifically for bluffing in TCs: an opponent who
receives a face-down offer of three cards cannot distinguish an offer worth 150 coins from one worth zero. This information
asymmetry is the foundation of the game's deception mechanics.

On each turn, the active player draws and auctions a card. The auctioneer either accepts the
highest bid or exercises a \emph{buy-right} (paying that amount to keep the card); with no bids,
they get it free. This creates a dilemma: high bids enrich the auctioneer, low bids hand them the
card cheaply. Overbidding one's holdings reveals wealth to all players and restarts the auction.

Instead of auctioning, the active player may initiate a TC with any opponent who shares
an animal type. The initiator makes a face-down offer using any money cards, including 0 bluffs. The
opponent may accept (taking the money sight-unseen and surrendering their animals) or counter with
their own face-down offer. Both offers are then revealed and the money is exchanged: each player
receives the other's coins, and the higher total wins. If both players hold a pair (2 or more)
of the animal, the loser transfers two contested cards to the winner; otherwise the loser
transfers one. Bluffing succeeds only
if the opponent accepts without seeing the offer; any positive counter defeats a 0 bluff. This
mechanism is the strategic core of the game, requiring calibrated risk assessment: overpaying wastes
resources, but underbidding loses cards.

Donkey cards distribute coins to all players on draw (50, 100, 200, 500 for the first through
fourth donkey), injecting liquidity and creating phase transitions: early games are
cash-constrained; late games permit larger bids.

\subsection{Benchmark Design}
\label{sec:design}

The benchmark consists of a game engine, an LLM agent framework, and an evaluation system. We
implement the complete \textsc{Cattle Trade} rules in Python with three auction modes: canonical
(simultaneous-intent bidding faithful to tabletop play), fast (single sealed bid), and legacy
(sequential). The engine validates all actions, handles exact money card accounting, and logs
complete state transitions.

Agents receive natural language observations (hand, visible cards, money, legal actions) and respond
with a structured JSON action specifying exact bid amounts for auctions and exact card compositions
for TC offers (e.g., \texttt{offer\_cards: [50, 10, 0]}), giving models direct control over
bluff construction rather than delegating card selection to a heuristic.
Each agent maintains a personal scratchpad updated each turn via an additional model call,
and receives the same system prompt (the \texttt{optimal} character of
Appendix~\ref{app:prompts}: neutral game rules plus ``play optimally to maximize your expected
score''), with no strategy hints or tactical guidance.
We use TrueSkill~\citep{herbrich2006trueskill} for competitive rating, which provides Bayesian
skill estimates ($\mu$, $\sigma$).

The framework handles malformed actions via multi-stage JSON fallback and retries truncated
responses. In canonical mode, the engine permits overbids by design and applies the tabletop penalty
(wealth revelation and auction restart) rather than silently correcting, preserving this as an
observable strategic error. When multiple bidders submit equal highest bids in the same call round,
ties are broken by a per-auction random priority order, sampled once when the auction starts.
Auction settlement uses a dynamic-programming subset-sum routine to construct the minimum-overpay
payment from the winner's money cards (no change is given); TC offers and counters, by
contrast, are composed card-by-card by the agent, so bluff construction (including 0-value cards) is
a deliberate choice rather than an engine heuristic. Full architecture details are in
Appendix~\ref{app:agent_spec}, and prompts are in Appendix~\ref{app:prompts}.

\subsection{Scalability and Code Agents}
\label{sec:scalability}

The benchmark supports parameterized configurations: set size $k \in \{2,3,4,5\}$ (cards per animal
type), animal types (1--10), player count (2--5), and starting hands (0--$k$). These combine into
presets from micro-duel (2 players, $k{=}2$, 3 types, ${\sim}$20 turns, ${\sim}$50 LLM calls) to
marathon (5 players, $k{=}5$, 10 types). Our experiments use the standard configuration (4 players,
$k{=}4$, 10 types).

We provide three deterministic code agents as calibration baselines, each implementing a distinct
strategic heuristic with no learning:
\begin{enumerate}
\item \textsc{TrackerAgent} maintains perfect information from all observable events (tracking
    revealed cards, opponent holdings, and estimated wealth) and makes greedy decisions conditioned on
    this state. It bids just above an opponent's estimated budget, exercises buy-right only for
    quartet-completing cards, and adjusts TC offers to the target's inferred cash position.
\item \textsc{SetRaceAgent} greedily pursues quartet completion, bidding aggressively on
    near-complete sets and ignoring animals with no path to a quartet. It evaluates each auction card by
    marginal gain toward a quartet and prioritizes TCs that complete or advance sets,
    regardless of cost.
\item \textsc{EconomyAgent} models the game as a resource economy, tracking money flows and donkey
    payouts to exploit cash-poor opponents and avoid enriching cash-rich ones. It caps spending at a
    fraction of its total wealth per turn, avoids buy-right when the auctioneer surplus would benefit a
    leading opponent, and bluffs only against opponents with low counter probability.
\end{enumerate}

% Experiments — setup, main results, strategic behavior, game-state adaptation

\section{Experiments}

\subsection{Setup}

We evaluate seven cost-efficient LLMs spanning four providers: Claude Haiku~4.5 (\textsc{Haiku},
Anthropic), Claude Sonnet~4.5 (\textsc{Sonnet}, Anthropic),
Gemini 2.5 Flash Lite (\textsc{G2.5-FL},
Google), Gemini~3 Flash Preview (\textsc{G3-F}, Google), Gemini~3.1 Flash Lite Preview
(\textsc{G3.1-FL}, Google), GPT-5.4 Nano (\textsc{GPT5.4-N}, OpenAI), and DeepSeek~v3.2
(\textsc{DS-v3.2}, DeepSeek). We pit each LLM against three deterministic code agents
(\textsc{TrackerAgent}, \textsc{SetRaceAgent}, \textsc{EconomyAgent}) to calibrate LLM play against
algorithmic baselines (Section~\ref{sec:code_agents}).

We run 228 primary games across two formats: (i)~60 pure-LLM tournaments in 15 balanced four-player
combinations, covering every pair of the six primary LLMs at least twice, to capture head-to-head
dynamics; and (ii)~168 mixed-format games where a single LLM competes against three code agents
across seven opponent-composition schedules (e.g., Tracker+Economy+SetRace, denoted C1, or
2$\times$Tracker+Economy, denoted C2), to calibrate each LLM against fixed algorithmic baselines. An
additional 14 exploratory games extend coverage to Sonnet~4.5 (10 pure-LLM matchups balancing
opponents across the field, plus 4 comp1 games against the code agents), yielding 242 games in
total. Sonnet results are reported with broader uncertainty ($\sigma{\approx}1.6$ vs.\
${\approx}1.0$ for the others) due to the smaller sample. All games use canonical auction mode
(simultaneous-intent bidding), full deck (10 animal types), full memory mode (scratchpad),
temperature~0.1, reasoning effort set to low, and a 4{,}096-token response limit. The system
prompt is the \texttt{optimal} character: the neutral game-rules block (Appendix~\ref{app:prompts})
followed by a single instruction to ``play optimally to maximize your expected score.'' No
strategy hints, examples, or tactical guidance are provided, so any strategic behaviour is
elicited from the model itself.

\subsection{Main Results}

\begin{figure}[tb]
\centering
\includegraphics[width=0.88\linewidth]{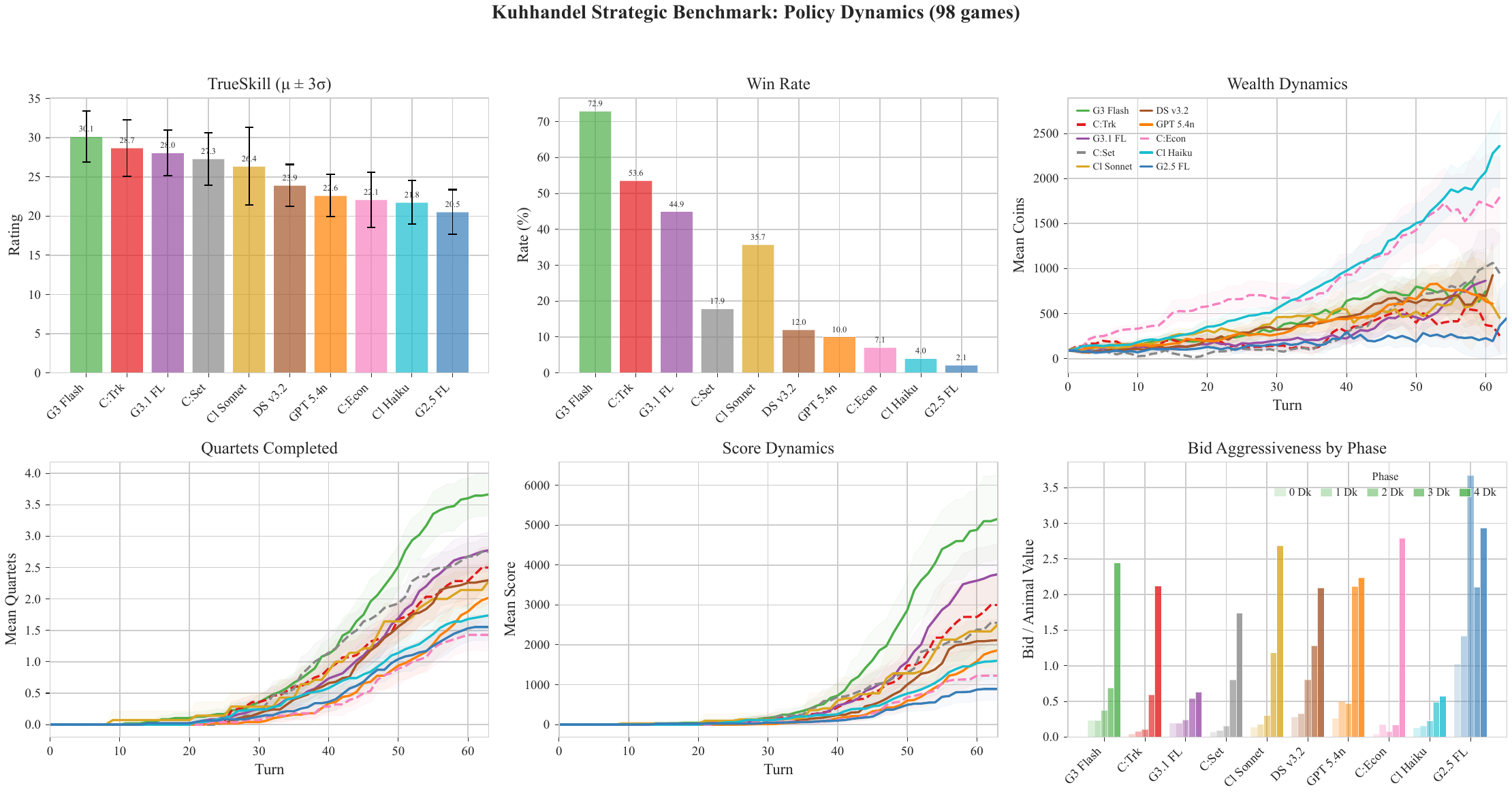}
\caption{\textbf{Tournament policy dynamics} (98 canonical games: 70 pure-LLM + 28 mixed
comp1). Agents ordered by TrueSkill posterior mean $\mu$; error bars on the top-left panel
show $\pm 3\sigma$ (99.7\% interval). Top row: TrueSkill, win rate, and per-agent mean
wealth across turns. Bottom row: per-agent mean quartets completed and mean score across
turns (each line is the mean over $\sim$50 games for that agent; shaded band is the
bootstrap 95\% CI on the mean), then bid aggressiveness by donkey phase. Quartets and score
are monotonically non-decreasing per game (with $k{=}4$ quartet permanence) and each game
is snapshotted at every turn with state carried forward between events, so the cross-game
mean is also strictly non-decreasing. Wealth (which can fluctuate) is reported only over games still running
at each turn; the surviving pool shrinks at late turns and is therefore biased toward
longer, more liquid games. Accept rate, bluff ratio, and challenger/target win rates are in
Appendix~\ref{app:detailed}.}
\label{fig:scores}
\end{figure}

Figure~\ref{fig:scores} summarises the two-format evaluation. On the combined-comp1 slice
(70 pure-LLM + 28 mixed-comp1 = 98 canonical games), we rank by TrueSkill $\mu$ with $\pm 3\sigma$
error bars (the lower edge is $\mu_c$). G3-F leads ($\mu{=}30.1\!\pm\!3.3$, 72.9\% wins, median
5{,}250), followed by TrackerAgent ($28.7\!\pm\!3.6$, 53.6\%), G3.1-FL ($28.0\!\pm\!2.9$, 44.9\%),
and SetRaceAgent ($27.3\!\pm\!3.3$): TrackerAgent beats six and SetRaceAgent five of seven LLMs.
G3.1-FL has a slightly higher median (3{,}930) than TrackerAgent (3{,}550) despite a lower
$\mu$: TrueSkill rewards how often an agent finishes ahead, so TrackerAgent's higher win
rate (53.6\% vs.\ 44.9\%) wins on $\mu$, while G3.1-FL's runner-up finishes carry larger
score tails. Sonnet~4.5 sits mid-field ($26.4\!\pm\!4.9$, $n{=}14$, 35.7\% wins; the wider
band reflects sample size, not weak play). The weak tier: DS-v3.2 ($23.9\!\pm\!2.7$), GPT5.4-N ($22.6\!\pm\!2.7$), EconomyAgent
($22.1\!\pm\!3.5$), Haiku ($21.8\!\pm\!2.8$), G2.5-FL ($20.5\!\pm\!2.9$). The score gap
tracks quartet completion: G3-F averages 3.96 quartets per game, Haiku and EconomyAgent below~1.7.

\subsection{LLM vs.\ Code-Agent Calibration}
\label{sec:code_agents}

\begin{figure}[tb]
\centering
\begin{minipage}[b]{0.48\linewidth}
\centering
\includegraphics[width=\linewidth]{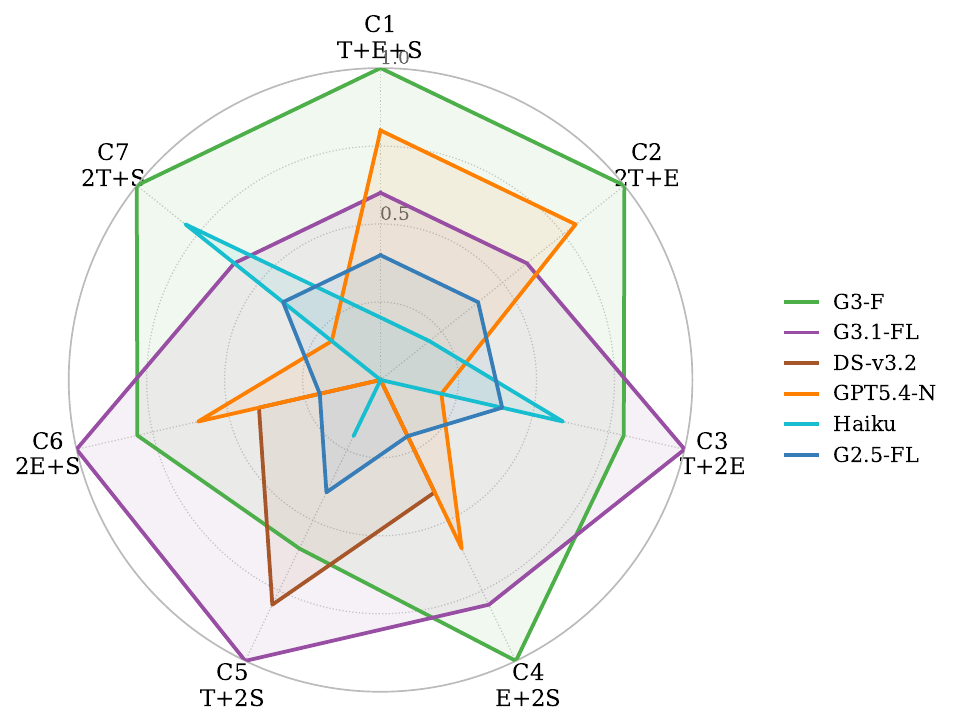}
\end{minipage}\hfill
\begin{minipage}[b]{0.48\linewidth}
\centering
\includegraphics[width=\linewidth]{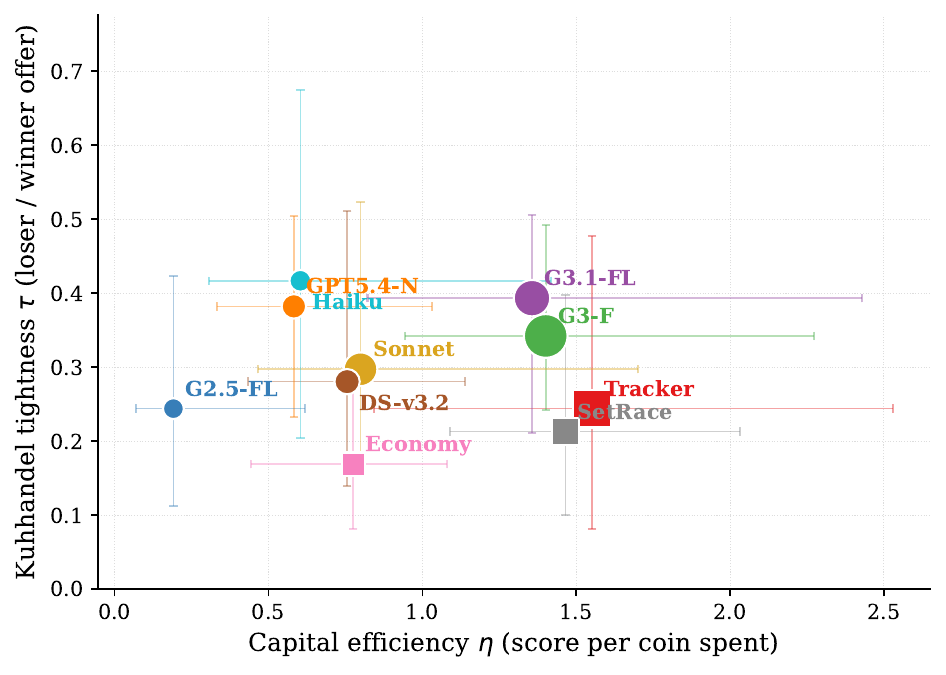}
\end{minipage}
\caption{\textbf{(a) LLM robustness across 7 code-agent compositions} (exp2\_all7, 168 mixed
games, 4 games per (LLM, composition) cell; Sonnet's 4 comp1 games omitted due to partial
coverage). Axes C1--C7 are compositions; values are per-composition win-rate percentiles
across the LLMs. Outward $=$ beats more peers on
that mix; large convex polygon $=$ robust across mixes, spiky polygon $=$ composition-sensitive.
\textbf{(b) Economic profile} (98 canonical games). $x$: capital efficiency $\eta$, $y$:
TC tightness $\tau$ (definitions in Appendix~\ref{app:metrics}). Median across games, IQR
as bars; circles $=$ LLM, squares $=$ code agent; marker size $\propto$ win rate.}
\label{fig:main_pair}
\end{figure}

The mixed-format games let us separate LLM skill from LLM-vs-LLM variance. In the 172-game exp2
slice (Figure~\ref{fig:main_pair}a; 168 core + 4 Sonnet comp1), G3-F wins 67.9\% of its 28 games
averaged over the seven compositions, and G3.1-FL wins 50.0\%. Both clear all three heuristic
baselines. The remaining LLMs fall below every code agent by mean score: DS-v3.2 (10.7\% wins),
GPT5.4-N (14.3\%), Haiku (7.1\%), and G2.5-FL (0.0\%). Sonnet won 1 of its 4 comp1 games
(25\%); its small sample prevents a per-composition robustness plot, so it is omitted from the
radar in Figure~\ref{fig:main_pair}a. The hardest compositions
for LLMs are those containing two TrackerAgents (C2, C7): the card-counting pressure compounds
and only G3-F still wins a majority. Conversely, compositions with two EconomyAgents (C3, C6) are
the softest: every LLM except G2.5-FL wins a larger share there, confirming that EconomyAgent's
conservative budgeting is the weakest of the three code heuristics.

TrackerAgent ($\mu=30.8$ in the 172-game slice) exploits observed events via card counting, a
capability no LLM replicates despite receiving the same observable information in its prompt.
SetRaceAgent ($\mu=29.7$) targets quartet completion aggressively, and EconomyAgent
($\mu=26.4$) relies on conservative resource management. The ordering TrackerAgent $>$ SetRaceAgent $>$ EconomyAgent suggests that information exploitation and decisive acquisition matter more than caution.
The economic profile in Figure~\ref{fig:main_pair}b complements this on two rates
(full definitions in Appendix~\ref{app:metrics}). The $x$-axis is capital efficiency $\eta{=}\text{score}/\text{gross outflow}$
(median across games); $\eta{>}1$ means the agent extracts more than one score point per coin
spent (possible because free acquisitions, won bluffs, and the multiplicative scoring
formula amplify returns). Further right $=$ each coin bought more final score. The $y$-axis is
TC bargaining tightness
$\tau = \sum_i (o^i_{\text{loser}} + 10)\,/\,\sum_i o^i_{\text{winner}}$,
size-weighted over counter-exchange wins. Money cards step by 10, so
$o_{\text{loser}}+10$ is the minimum bid the winner needed to win;
$\tau{=}1$ $=$ the winner paid only that minimum increment, while
$\tau{\to}0$ $=$ the winner crushed a small counter with a huge
offer (loose, most coins wasted). Further up $=$ fewer coins burned per TC won. The
ideal corner is top-right: efficient \emph{and} tight. G3-F sits closest, combining the
highest $\eta$ among LLMs with solid $\tau \approx 0.34$. Haiku and GPT5.4-N are the tightest
bargainers ($\tau \approx 0.4$) but pay for it with low efficiency. Tracker and SetRace land
opposite: high efficiency but looser counters ($\tau \approx 0.2$--$0.25$), trading
bargaining precision for acquisition pressure.

\subsection{Strategic Behavior}

Three mechanisms separate strong from weak play (full metrics in Table~\ref{tab:behavior}).

\textbf{Spending efficiency.}
Capital efficiency $\eta$ (median of per-game score$/$gross outflow; see
Appendix~\ref{app:metrics}) places G3-F at $\eta{=}1.77$ and G3.1-FL at 1.46; TrackerAgent
reaches 1.55. Values above 1.0 are possible because free acquisitions (no-bid auctions), won
bluffs, and the multiplicative scoring formula amplify returns beyond one-to-one.
G2.5-FL spends aggressively (bid aggressiveness 2.52) but only extracts $\eta{=}0.23$,
overpaying for cards it fails to assemble into quartets. High volume without
targeting yields worse outcomes than moderate volume with better targeting. Cost per quartet
(Appendix~Figure~\ref{fig:cost_per_quartet}) traces the same ordering: top agents pay
600--750 coins per quartet overall; G2.5-FL pays 1{,}193.

\textbf{Resource discipline and self-bidding.}
The three code agents never overbid. G2.5-FL overbids at 1.20\%, the highest rate, and finishes
last; Haiku (0.87\%), DS-v3.2 (0.49\%), and GPT5.4-N (0.47\%) also overbid at meaningful rates.
Among LLMs, self-bidding (rounds where the agent raises with no competing bid since its last
bid) correlates inversely with performance: G3.1-FL self-bids in under~7\% of rounds while
DS-v3.2, GPT5.4-N, and G2.5-FL self-bid in over~74\%. We cannot tell from the rate alone
whether this reflects a failure to detect non-competitive contexts or a deliberate
escalation when an agent really wants the card; one DS-v3.2 trace incrementing 10$\to$850
over 49 sole-bidder rounds is suggestive of the former.

\textbf{Buy-right and phase adaptation.}
G3.1-FL and G3-F exercise buy-right most frequently among LLMs (31.3\% and 26.5\%); TrackerAgent
tops all agents at 34.4\%. G3-F escalates bid intensity from early-game (0.26) to late game
(2.49), an $\approx$10$\times$ ramp timed to quartet-completion pressure. G3.1-FL shows a
smaller, coherent ramp (0.22 $\to$ 0.65). G2.5-FL stays at bid aggressiveness 2.07 early
\emph{and} 2.08 late, with no adaptation, skipping the accumulation phase entirely. DS-v3.2 and
GPT5.4-N escalate to similar absolute values (1.65 and 2.09) but without the spending discipline
to convert aggression into quartets. The code agents themselves ramp sharply (Tracker
0.06$\to$1.92, SetRace 0.11$\to$1.55), so endgame escalation is a \emph{rational} pattern the
top LLMs have learned and the weakest ones have not.

\subsection{Trajectory-Level Observations}

Aggregate metrics mask qualitative behaviour visible in individual traces.
G2.5-FL repeatedly initiates TCs after depleting its
money through overbidding; in one game it challenges for a goose with zero money cards
(producing a 0-value offer the opponent beats with a 10-coin counter for free), failing to
condition action choice on resource state.
G3-F conditions TC offers on both opponent wealth
and game context, deploying 0-value bluffs against bankrupt opponents and 500+\,coin offers
against well-funded ones contesting high-value animals; one trace shows it completing a
fourth quartet by paying far above face value, with the multiplicative-scoring formula
turning a nominally wasteful overpay into a net gain of $\approx 1{,}800$ points. Weaker LLMs
do not exhibit this portfolio-aware price discrimination.
Sole-bidder runs (a single bidder raising while others
pass) appear in both LLMs and code agents at high rates because canonical mode requires all
bidders to pass \emph{in the same round} for the auction to terminate. DS-v3.2 incrementing
from 10 to 850 over 49 call rounds on a sole-bidder run is suggestive of the
``treats-own-bid-as-competitor'' pathology, but a single trace cannot distinguish that from
a deliberate escalation when the bidder really wants the card or expects an opponent to.
The 75.4\% self-bidding rate (Table~\ref{tab:behavior}) is therefore best read as a
within-auction raising propensity, not a one-sided failure mode.

\subsection{Computational Cost and Format Compliance}

Token usage varies roughly 20$\times$ across models: G3-F emits around 1{,}500 completion tokens
per LLM call on average (reflecting extended internal reasoning), G3.1-FL about 80. Both rank in the
top two, so verbose reasoning chains are not required for strong play. Haiku emits the most
completion tokens but ranks ninth: token volume alone does not predict strategic quality.
Structured-output failures (invalid JSON, missing fields, truncation) are below 1\% for every
model we evaluated; format errors trigger a single retry with explicit error feedback and do not
explain the performance gaps we observe (Appendix~\ref{app:detailed}).

% Discussion
\section{Discussion}

Top LLMs avoid systematic failures (overbidding,
self-bidding, bankrupt initiation) that plague weaker models and never surface in code agents:
conditional logic already suffices where LLMs still fail. Strategic coherence in turn (spending efficiency, resource discipline, adaptive phase play) is associated with success. 

TrackerAgent's second-place ranking ($\mu=28.7$, above six of seven LLMs) calibrates the
benchmark: card-counting heuristics suffice to outperform most LLMs tested. The code-agent ordering
(TrackerAgent $>$ SetRaceAgent $>$ EconomyAgent) shows that information exploitation matters more
than greedy quartet-chasing, which in turn outperforms conservative budgeting. This mirrors the LLM
ranking, where the top models combine active acquisition with resource discipline, while weaker
models either spend recklessly or participate too passively.

Our ranking by posterior mean (G3-F $\gg$ G3.1-FL $>$ Sonnet~4.5 $>$ DS-v3.2 $>$ GPT5.4-N $>$
Haiku $>$ G2.5-FL, with Sonnet still wider in CI at $n{=}14$) broadly aligns with Chatbot
Arena~\citep{lmarena2026,zheng2024judging}\footnote{May 2026 snapshot Elo: G3-F 1474, Sonnet
1453, G3.1-FL 1439, DS-v3.2 1423, Haiku 1408, GPT5.4-N 1406, G2.5-FL 1380.} on the top and
bottom of the field, with two within-tier swaps in the middle (G3.1-FL/Sonnet, GPT5.4-N/Haiku).
It diverges more sharply from reasoning-mode-aggregating evaluations such as the Artificial
Analysis Intelligence Index v4.0~\citep{artificialanalysis2026}\footnote{Index v4.0: GPT5.4-N
(xhigh) 44, Sonnet 43, DS-v3.2 42, Haiku (reasoning) 37, G3.1-FL 34, G3-F (reasoning) 46.},
where reasoning-aware scoring reorders cells that rank in the opposite direction in our
setting (most starkly G3.1-FL: second in ours, last in the Index). Multi-turn
strategic play depends on capabilities (state tracking, adaptive resource allocation,
structured-output reliability) that static benchmarks do not measure but that conversational
evaluations partially capture.

The benchmark's diagnostic value lies in identifying why a model loses, not just that it loses.
Overbid frequency, self-bidding rate, bankrupt-initiation patterns, and context-dependent offer
calibration are failure modes invisible to both static evaluations and aggregate rankings like Elo.
The structured game logs make these behaviours directly observable and quantifiable.

Limitations: all seven LLMs are cost-efficient models at low reasoning effort and a
4{,}096-token budget; Sonnet~4.5 has only $n{=}14$ games. We test only full-memory mode and
do not ablate memory size or temperature, and behavioural metrics are point estimates without
confidence intervals. Within-agent score std (e.g., G3-F 4{,}026 on median 5{,}250) dominates
the cross-seat win-rate differentials (Appendix~\ref{app:detailed}), suggesting deck order
matters more than seat position; a controlled ablation is future work. Whether these failure
modes generalise to other economic settings remains untested.

Some failures may reflect prompt design rather than model limitations. All models receive identical
prompts describing game rules and legal actions but we do not explicitly instruct them to track
wealth before bidding or check whether other players are bidding before raising. The code agents
receive no natural-language prompts at all and still avoid these errors through conditional logic,
suggesting the underlying issue is one of reasoning rather than instruction-following. Code agents
also operate on structured data with exact arithmetic, while LLMs must parse natural-language
observations and track state across turns; some failures classified as reasoning errors (e.g.,\
overbidding) may partly reflect numerical parsing or working-memory limitations.

% Conclusion
\section{Conclusion}

We introduced \textsc{Cattle Trade}, a multi-agent benchmark integrating auctions,
hidden-offer TCs, bluffing, and resource management. Across 242 games, strategic
coherence (spending efficiency, resource discipline, adaptive phase play) is associated with
success more strongly than any single capability. Two of the three deterministic heuristics
outperform most tested LLMs, suggesting cost-efficient models lack not individual skills but
their reliable integration under competitive pressure. The structured logs pinpoint
\emph{why} models fail (overbidding, self-bidding spirals, bankrupt initiation), failure
modes invisible to aggregate evaluations and pointing to concrete gaps in state tracking,
resource-constraint reasoning, and opponent modelling that static benchmarks do not measure. More broadly, benchmarks of this kind may become increasingly important for evaluating agents beyond static reasoning or single-turn task success. They make it possible to test whether models can sustain strategic coherence over time, manage resource constraints, and adapt interactively in multi-agent environments with conflicting incentives. This matters for prospective deployments in economic and organizational settings, where failures often arise not from the absence of any single capability, but from the inability to coordinate multiple capabilities across extended interactions. In this sense, \textsc{Cattle Trade} is a step toward evaluating agentic competence under more realistic conditions of strategic interaction.

Several directions follow. Evaluating frontier models at full reasoning budgets would test
whether the failures we document reflect fundamental limitations or a cost-efficiency
tradeoff of smaller models. Enabling inter-agent chat would introduce the meta-communication
layer present in human play (coalition signalling, persuasion, relationship management) that
the current structured-action format excludes. Self-play fine-tuning of open-weight models
could test whether strategic failure modes such as self-bidding are trainable away. Finally,
benchmarking against human players would calibrate whether LLM failures reflect genuinely
hard strategic problems or errors that novice humans also avoid.

\bibliography{bib/refs}
\bibliographystyle{iclr2026_conference}

\appendix
% Appendix A — Game rules summary
\section{Game Rules Summary}
\label{app:rules}

Table~\ref{tab:animals} lists all animal types and their quartet values. The game uses 10 animal
types with 4 cards each (40 cards total).

\begin{table}[h]
\caption{Animal types and quartet values in \textsc{Cattle Trade}.}
\label{tab:animals}
\vspace{0.5em}
\centering
\small
\begin{tabular}{lr}
\toprule
\textbf{Animal} & \textbf{Quartet Value (points)} \\
\midrule
Chicken & 10 \\
Goose   & 40 \\
Cat     & 90 \\
Dog     & 160 \\
Sheep   & 250 \\
Goat    & 350 \\
Donkey  & 500 \\
Pig     & 650 \\
Cow     & 800 \\
Horse   & 1{,}000 \\
\bottomrule
\end{tabular}
\end{table}

\paragraph{Auction flow.} The active player draws a card and announces an auction. Canonical
mode runs as iterative \emph{call rounds}: in each round, all non-auctioneer (non-eliminated)
players submit a bid intent (a number or \emph{pass}) simultaneously. The highest valid bid in
the round becomes the current standing bid (ties broken by a per-auction random priority order;
see Appendix~\ref{app:agent_spec}). The auction continues into further rounds (bidders may
raise across rounds) and ends the first time all eligible bidders pass while a standing bid
exists. If \emph{no} bid is placed in round~0, the engine runs one confirmation round; if that
round is also empty, the auctioneer receives the card for free. This two-round rule prevents a
single miscoordination on the blind first call from instantly handing the card to the
auctioneer. Once the auction ends, the auctioneer sees the winning bid and decides: accept
(card goes to bidder, money to auctioneer) or buy-right (auctioneer pays the bid amount to the
bidder, keeps the card). If a player bids more than they have, their wealth is revealed to all
players, the auction is repeated, and the penalized player is limited to bidding their true
valid amount.

\paragraph{Trade challenge flow.} The active player selects an opponent who shares at least one animal
type. The initiator places money cards face-down as an offer. The opponent chooses: (a)
\emph{accept:} take the money (without seeing it) and give cards of the contested animal to the
initiator; or (b) \emph{counter:} place their own face-down offer. If countered, both offers are
revealed and exchanged: each player receives the other's offer. The player who \emph{offered
more money} wins. If both sides hold a pair (2 or more) of the animal, the loser transfers two
contested cards to the winner; otherwise the loser transfers one.
Bluffing (0 offers) only works when the opponent
\emph{accepts} without seeing; if they counter with any positive amount, they beat the bluff. The
initiator's advantage is choosing when to attack and winning after 3 ties. On ties (equal amounts),
the TC repeats; after 3 ties, the initiator wins by default.

\paragraph{Donkey payout.} When a donkey card is drawn for auction, all players immediately receive
a money card from the bank (50, 100, 200, or 500 depending on which donkey card). This injects
liquidity into the game economy, changing bidding dynamics.

% Appendix B — Extended results, cross-play analysis, economics
\section{Detailed Results}
\label{app:detailed}

\subsection{Metric Definitions}
\label{app:metrics}

All per-agent metrics referenced in §4 and used in the figures and tables of this appendix are
computed from raw game logs as follows. TC abbreviates trade challenge. $\mathcal{Q}$ denotes
the set of completed quartets and $v_q$ the quartet value (§3.1).

\begin{description}
\item[Final score $S$.] $S = \big(\sum_{q\in\mathcal{Q}} v_q\big)\cdot\lvert\mathcal{Q}\rvert$
  (Eq.~\ref{eq:scoring}); evaluated at end of game.
\item[Capital efficiency $\eta$ ($\equiv$ \emph{Spend Eff.} in Table~\ref{tab:behavior}).]
  $\eta = S / \mathrm{out}$ per game, where $\mathrm{out}$ is gross outflow: the total coins
  leaving the agent's hand, summed across auction payments (when winning a bid), TC
  offers paid in accepted trades and counter-exchanges, and buy-right payments. Reported as
  median across games.
\item[TC bargaining tightness $\tau$.]
  For each counter-exchange won by the agent in a game, let $w_k$ be the winner's offer and
  $\ell_k$ the loser's. Then $\tau = \sum_k (\ell_k + 10) / \sum_k w_k$, size-weighted per
  game (summing over counter-wins $k$). Money cards step by $10$, so $\ell_k + 10$ is the
  minimum bid the winner needed to win; $\tau$ is bounded by $1$ because
  $w_k \geq \ell_k + 10$. $\tau{=}1$ $=$ winner paid only the minimum increment;
  $\tau{\to}0$ $=$ winner crushed a small counter with a much larger offer.
\item[Bid aggressiveness.] Mean of $\text{bid}/v_q$ over all auctions in which the agent bid,
  where $v_q$ is the quartet value of the auctioned animal.
\item[Buy-right \%.] Fraction of auctioneer decisions in which the agent exercised buy-right
  (keeping the card) rather than selling to the highest bidder.
\item[TC-accept \%.] Fraction of TCs (agent as target) resolved by
  accepting the face-down offer rather than countering.
\item[Bluff \%.] Fraction of the agent's TC offers consisting purely of $0$-value money
  cards.
\item[Self-bid \%.] Fraction of the agent's auction bids placed in rounds with no competing
  bid since this agent's last bid (i.e., the agent is raising while no other bidder has acted
  in the interim). Equivalently in canonical mode, the agent was the current high bidder when
  it bid again.
\item[Overbid \%.] Fraction of auctions in which the agent submitted a bid exceeding its total
  money; triggers the canonical-mode wealth-revelation penalty (§3.2).
\item[Win rate.] Fraction of games in which the agent achieved the highest final score $S$.
\item[TrueSkill $\mu, \sigma$.] Bayesian skill estimate~\citep{herbrich2006trueskill} fit
  across all games in the slice. The paper ranks by posterior mean $\mu$; $\mu_c = \mu - 3\sigma$
  is reported as the lower edge of the 99.7\% interval where useful.
\end{description}

\begin{table}[h]
\centering
\resizebox{\linewidth}{!}{%
\begin{tabular}{lccccccc}
\toprule
Agent & Bid Agg. & Buy-Right & TC-Accept & Bluff\% & Self-Bid & Spend Eff. & Overbid\% \\
\midrule
Gemini 3 Flash        & 0.91 & 26.5\% & 36.3\% &  8.8\% & 21.7\% & 1.77 & 0.00\% \\
TrackerAgent          & 1.14 & 34.4\% &  0.0\% & 27.0\% & 58.7\% & 1.55 & 0.00\% \\
Gemini 3.1 Flash Lite & 0.36 & 31.3\% & 32.1\% &  7.0\% &  6.8\% & 1.46 & 0.35\% \\
SetRaceAgent          & 0.83 & 31.2\% & 34.8\% & 23.0\% & 67.0\% & 1.46 & 0.00\% \\
DeepSeek v3.2         & 1.03 & 18.2\% & 45.1\% & 17.0\% & 75.4\% & 0.70 & 0.49\% \\
GPT-5.4 Nano          & 1.44 & 27.6\% & 28.6\% & 15.2\% & 74.6\% & 0.74 & 0.47\% \\
Claude Haiku 4.5      & 0.31 & 20.0\% & 26.4\% & 15.6\% &  9.1\% & 0.69 & 0.87\% \\
Claude Sonnet 4.5  & 0.89 & 22.2\% & 44.3\% & 15.7\% &  4.8\% & 0.80 & 0.43\% \\
EconomyAgent          & 0.67 & 14.5\% & 14.8\% & 30.6\% & 59.6\% & 0.78 & 0.00\% \\
Gemini 2.5 Flash Lite & 2.52 & 15.9\% & 51.1\% & 25.7\% & 78.5\% & 0.23 & 1.20\% \\
\bottomrule
\end{tabular}}%
\caption{Strategic-behaviour profiles across 98 canonical games (combined-comp1 slice). Bid Agg.:
bid-to-value ratio. Buy-Right: fraction of auctioneer decisions keeping the card. TC-Accept:
TCs accepted. Bluff\%: pure 0-value offers. Self-Bid: bids in rounds with no
competing bid since this agent's last bid. Sole-bidder runs are common in canonical mode for
both LLMs and code agents because the engine requires all bidders to pass in the same round
to terminate; we cannot reliably distinguish a model that is mistakenly raising against
itself from one that is deliberately escalating because it really wants the card or expects
an opponent to bid. Self-Bid\% should therefore be read as a within-auction raising
propensity, not a one-sided failure-mode rate. Spend Eff.: median per-game points per coin
spent. Overbid\%: overbid frequency. Sonnet is based on $n{=}14$ exploratory games (vs.\
47--50 for the other LLMs); its values carry wider uncertainty.}
\label{tab:behavior}
\end{table}

Score variance is high: G3-F ranges from 0 to over 10{,}000 (mean 6{,}232, std 4{,}026), reflecting
the multiplicative scoring formula's sensitivity to quartet count. G2.5-FL shows the highest
relative variance (std 1{,}283 on mean 1{,}060). Within-agent score std exceeds cross-seat
win-rate differentials by 1--2 orders of magnitude, indicating that deck-order variance dominates
seat-position variance as a driver of per-game outcomes.

\begin{figure}[H]
\centering
\includegraphics[width=\linewidth]{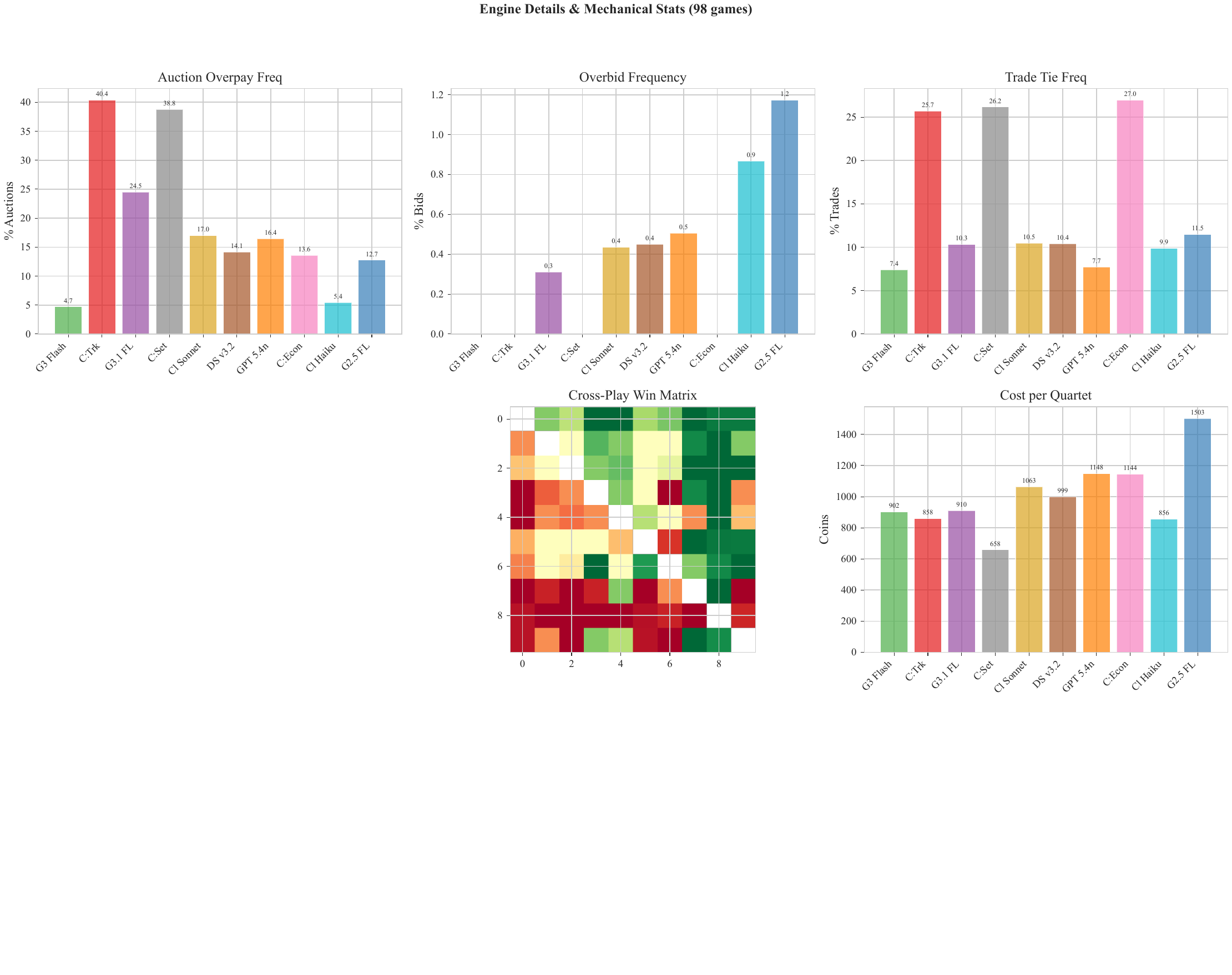}
\caption{Engine-level metrics (98 canonical games, 7 LLMs + 3 code agents): auction participation,
win rates, overpay/overbid frequencies, TC counter and challenger/target win rates, bluff
magnitude, tie frequency, token usage, truncation, and cost per quartet.}
\label{fig:appendix_metrics}
\end{figure}

Figure~\ref{fig:appendix_metrics} provides detailed auction and TC metrics across the ten
agents. The cross-play win-rate heatmap (bottom panel) shows G3-F winning the majority of games
against every opponent, including all three code agents. G3.1-FL maintains positive records against
all models except G3-F. TrackerAgent outperforms six of seven LLMs, while DS-v3.2 only beats
weaker LLMs.

\begin{figure}[H]
\centering
\includegraphics[width=\linewidth]{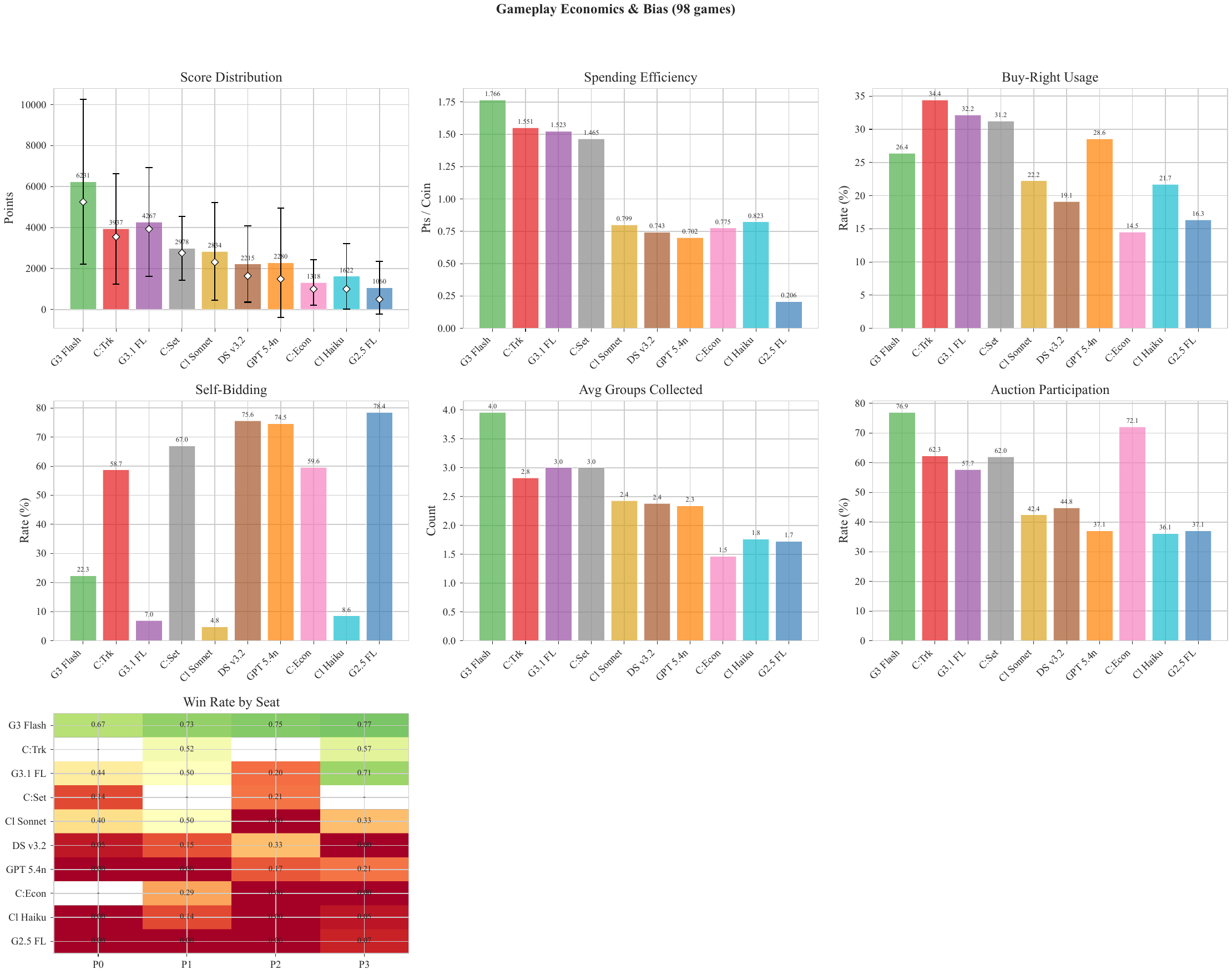}
\caption{Governance-level analysis (98 canonical games, 7 LLMs + 3 code agents): seat-position win
rates, money trajectories, accept rate by offer size, exploitation metrics, bid-to-wealth ratio,
and TC initiation by donkey phase.}
\label{fig:extended}
\end{figure}

Figure~\ref{fig:extended} shows positional analysis, economic trajectories, and phase-dependent
behaviour. The positional win-rate heatmap (left) confirms no systematic seat advantage for strong
models. Money trajectories (centre) show G3-F maintaining higher liquidity throughout games, while
G2.5-FL's cash reserves deplete rapidly under overbidding penalties. Code agents show stable money
trajectories under their deterministic rules. Wealth, unlike score, can fluctuate; we report
the per-turn mean only over games still running at that turn, so late-turn means are
biased toward longer-running, more liquid games. The
accept-rate-by-offer-size panel (right) shows
acceptance probability rising with the number of cards at stake, suggesting models calibrate
defensive decisions to trade magnitude.

\begin{figure}[H]
\centering
\includegraphics[width=0.75\linewidth]{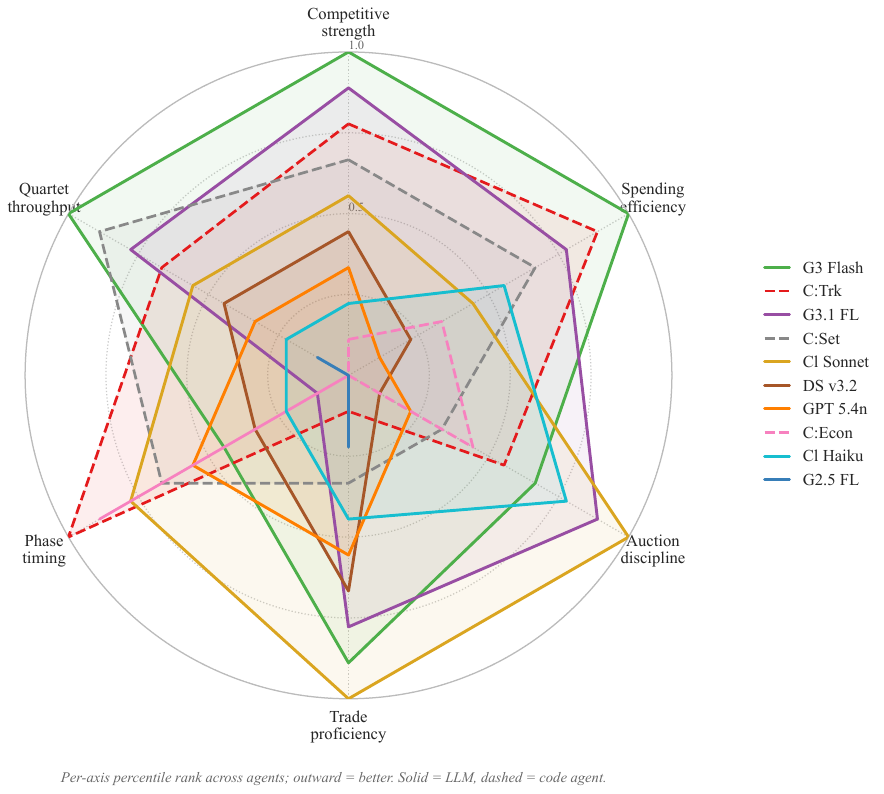}
\caption{\textbf{Strategic-profile radar} across six orthogonal axes (98 canonical games, all 10
agents). Each axis is percentile-normalised across the ten agents, so outward means
\emph{better}: competitive strength (TrueSkill $\mu$), spending efficiency (points per coin),
auction discipline ($1-\tfrac12(\text{overbid}+\text{self-bid})$), TC proficiency (TC
challenger win rate), phase timing ($\log$ late/early bid-aggressiveness ratio), and quartet
throughput (mean quartets per game). Solid $=$ LLM, dashed $=$ code agent.}
\label{fig:radar_strategy}
\end{figure}

\begin{figure}[H]
\centering
\includegraphics[width=\linewidth]{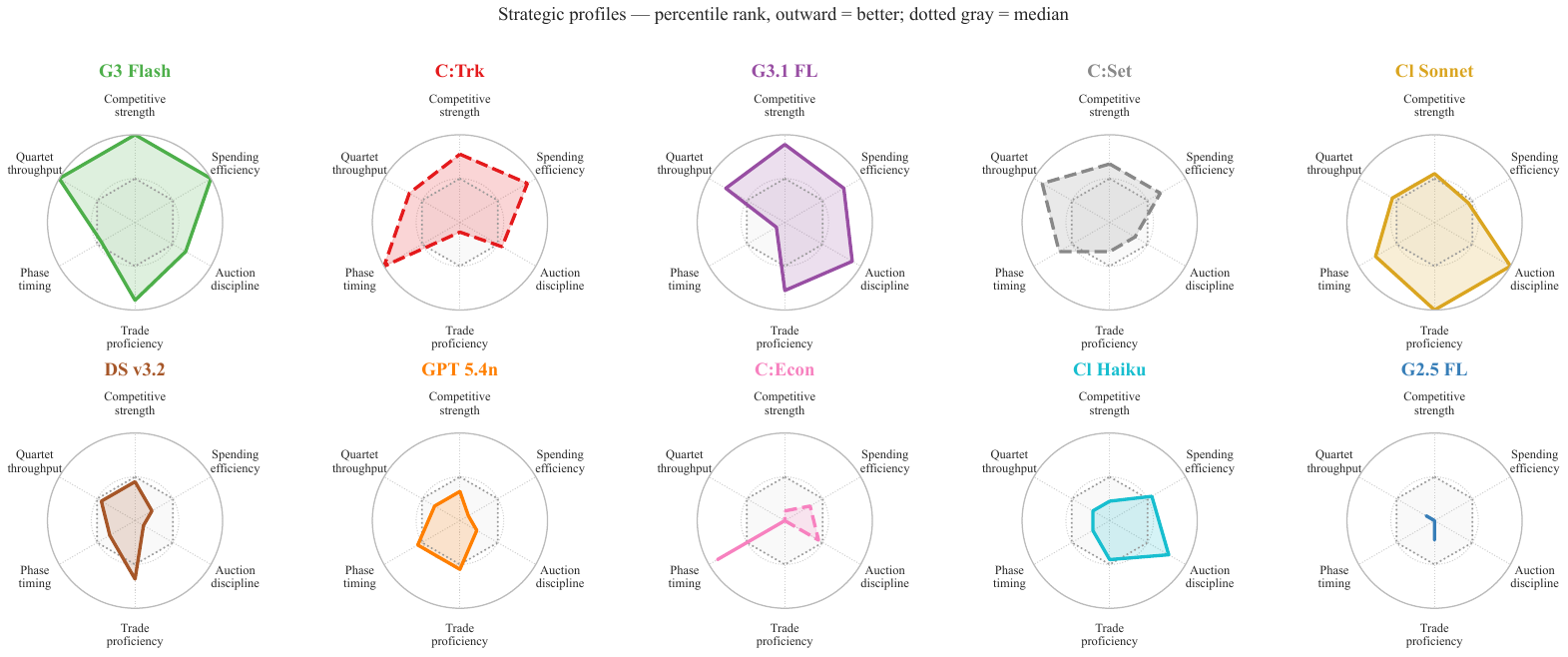}
\caption{Strategic-profile radar per agent (98 canonical games). Same six axes as
Figure~\ref{fig:radar_strategy}; each agent plotted against the cross-agent median (dotted grey).
The small-multiples view makes per-agent shape asymmetries easier to read than the overlaid
version.}
\label{fig:radar_small}
\end{figure}

\begin{figure}[H]
\centering
\includegraphics[width=\linewidth]{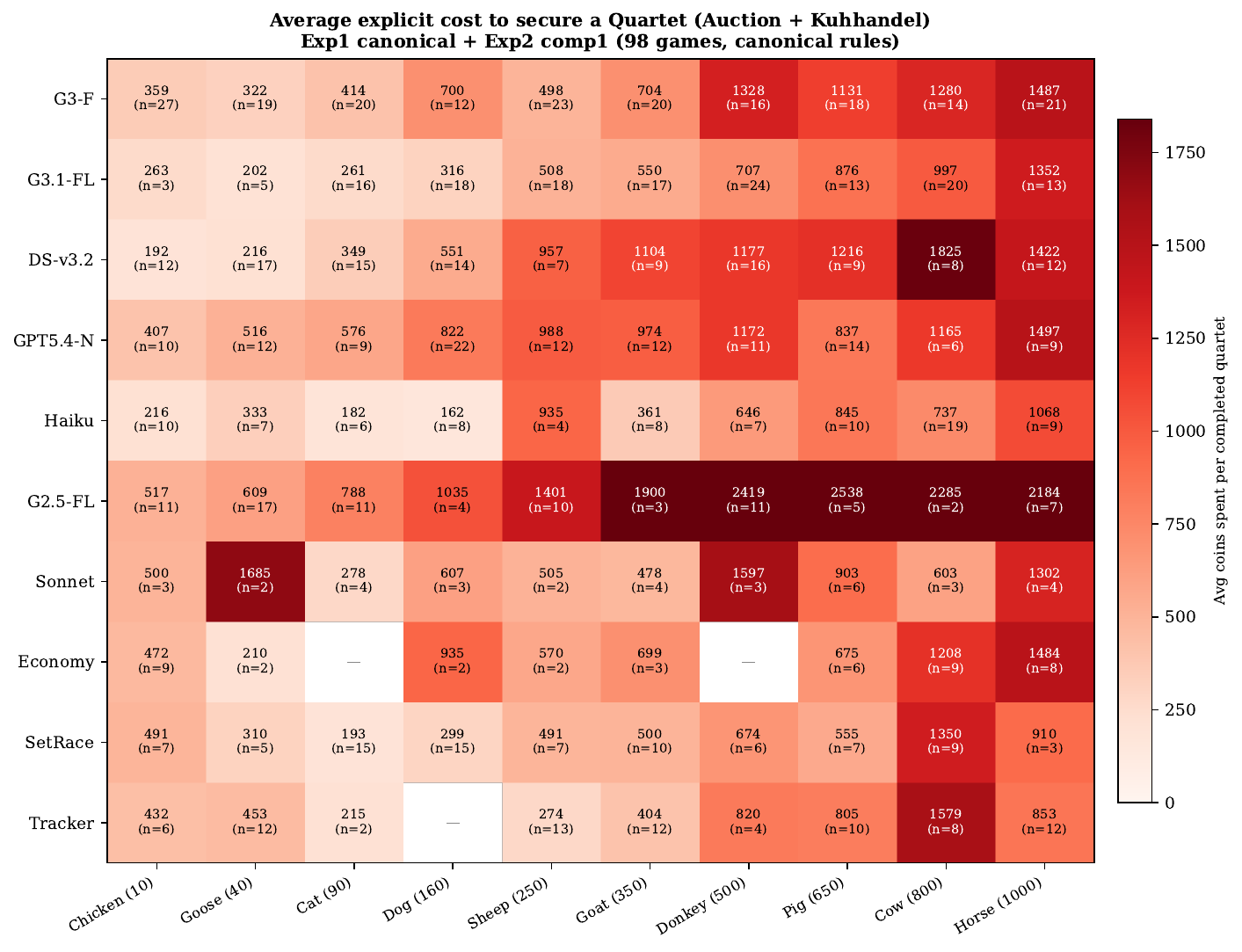}
\caption{\textbf{Gross-outflow cost per quartet} per animal (98 canonical games). Rows: 7 LLMs
then 3 code agents; columns: 10 animal types in ascending value order. Cell value: total coins
spent by that agent on that animal divided by quartets completed; $n$: quartets completed
(cells with $n{=}0$ are blanked, since the ratio is undefined). This
is a first-order view; it does not account for TC inflows or the multiplicative scoring
effect (see the net-ledger analysis in Figure~\ref{fig:main_pair}b).}
\label{fig:cost_per_quartet}
\end{figure}

\begin{figure}[H]
\centering
\includegraphics[width=0.95\linewidth]{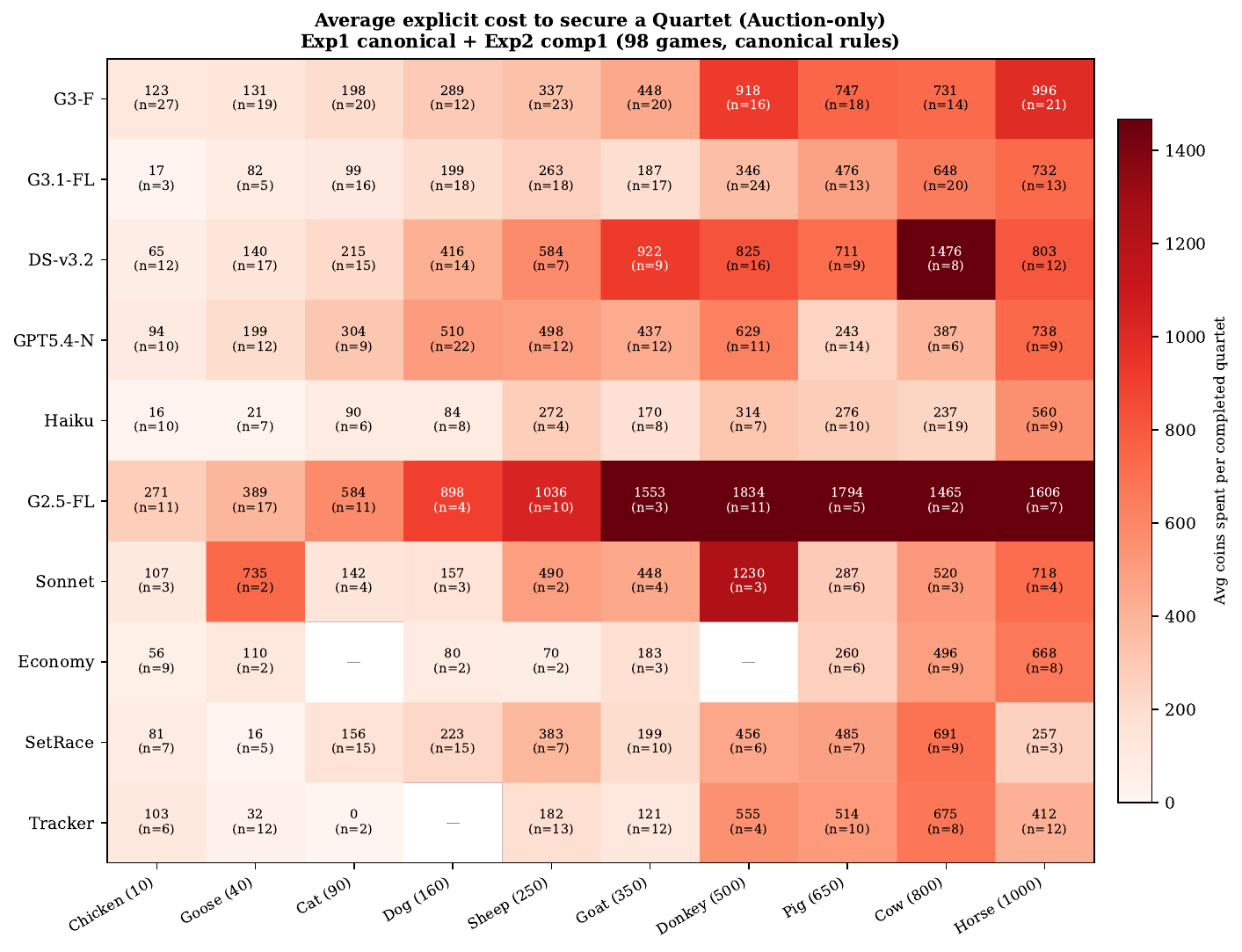}\\[0.3em]
\includegraphics[width=0.95\linewidth]{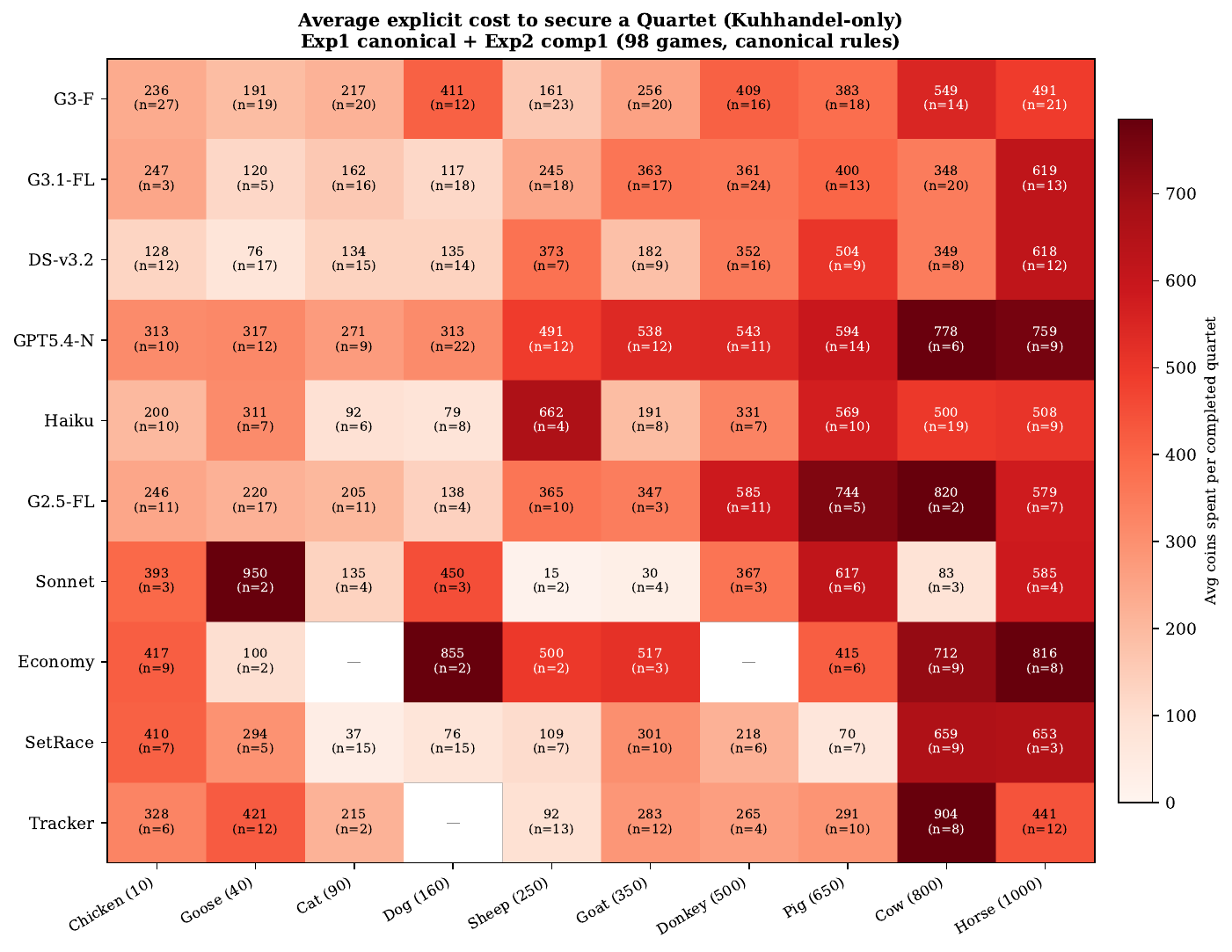}
\caption{Explicit cost decomposition per animal (98 canonical games). \textbf{Top:} auction-only
final prices paid by the winner. \textbf{Bottom:} TC-only winner's last-round offer value.
Auction costs dominate for high-value animals; TC costs concentrate on mid-tier animals.}
\label{fig:cost_per_quartet_decomp}
\end{figure}

\begin{figure}[H]
\centering
\includegraphics[width=\linewidth]{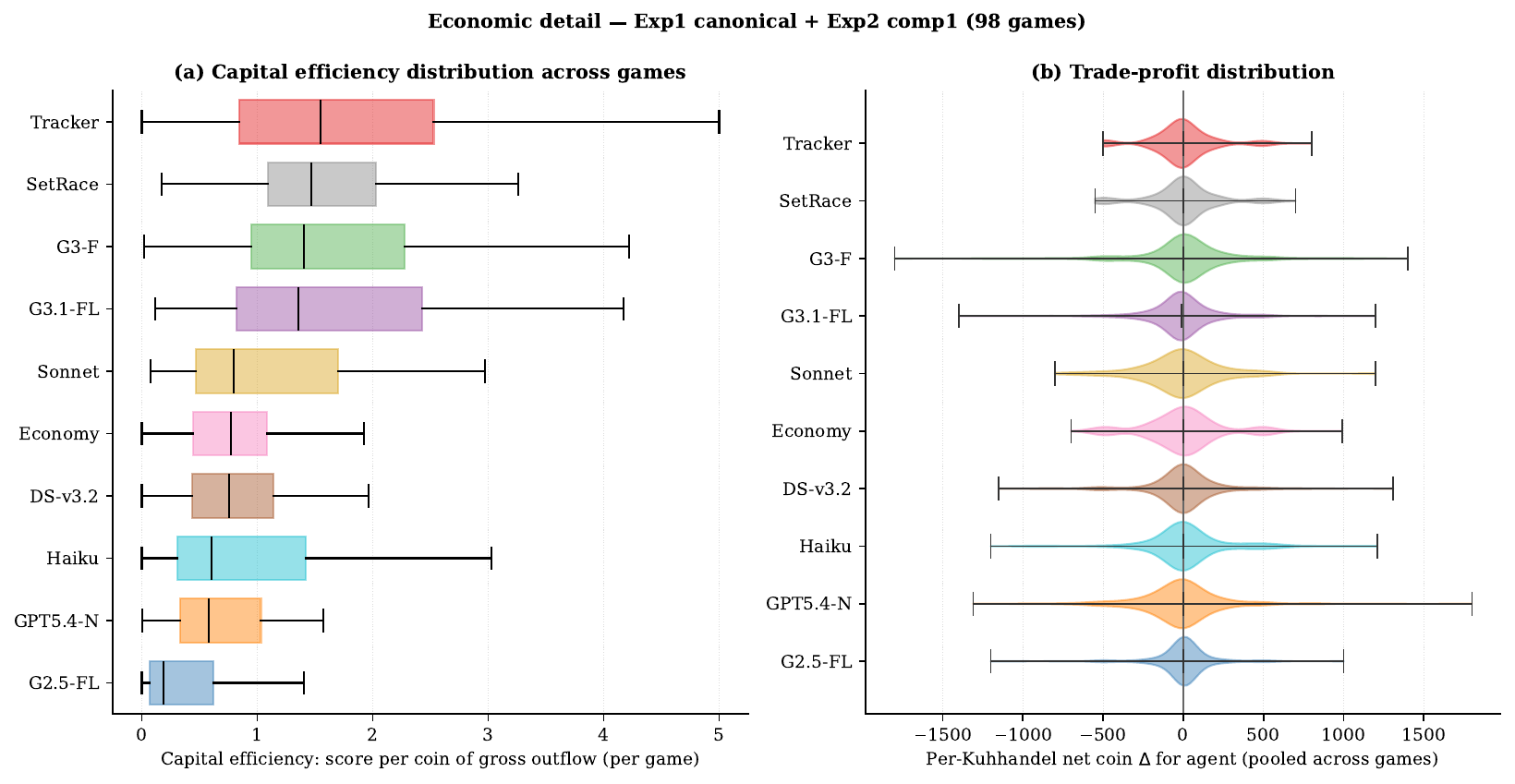}
\caption{\textbf{Economic detail} (98 canonical games). \textbf{(a)} Per-game capital efficiency
distribution, $\eta = \text{score}/\text{gross outflow}$. \textbf{(b)} Pooled per-TC net coin $\Delta$ per agent. Mass
right of zero $=$ agent pockets money on average; mass left $=$ agent pays to win animals in
TCs. The main-body scatter (Figure~\ref{fig:main_pair}b) aggregates these two marginals
into per-agent medians with IQR.}
\label{fig:economics_detail}
\end{figure}

Token usage per game varies roughly 20$\times$: G3-F generates about 275{,}000 completion tokens
per game, while G3.1-FL uses about 14{,}800. Both rank in the top tier, so verbose reasoning
chains are not required for strong strategic play in this setting.

% Appendix C — Agent observation/action interface, error handling, scratchpad
\section{Agent Architecture}
\label{app:agent_spec}

\paragraph{Observation space.}
At each decision point, player~$i$ receives an observation $o_t^i$ rendered as natural language. The
observation comprises four components:

\begin{enumerate}
\item \textbf{Private state}: exact money card denominations (e.g., 0$\times$2, 10$\times$4,
    50$\times$1) and animal holdings with counts.
\item \textbf{Public state}: all opponents' animal holdings (type and count), opponents' money card
    \emph{counts} (not values), number of cards remaining in the deck, and turn number.
\item \textbf{Decision context}: parameters specific to the pending decision, including the card
    being auctioned and current bids (auction), or the animal type, challenger/target identities, and
    tie count (TC).
\item \textbf{Memory} (if enabled): the most recent ${\sim}10$ game events (${\sim}$150 tokens) and
    the agent's scratchpad contents (see below).
\end{enumerate}

Information asymmetry is maintained: money card \emph{values} of other players are hidden; only card
counts are visible. A player's exact wealth is revealed only when they overbid in an auction,
mirroring the tabletop game's information structure.

\paragraph{Action space.}
The action space $\mathcal{A}_t$ is context-dependent. The engine presents a decision prompt and the
agent responds with a structured JSON object. TC abbreviates trade challenge. Seven action types arise:

\begin{enumerate}
\item \textbf{Turn choice}: \texttt{\{auction, kuhhandel\}}:select whether to draw a card for
    auction or initiate a TC with a valid target.
\item \textbf{Auction bid}: \texttt{\{bid(\emph{amount}), pass\}}:bid on the auctioned card or pass.
    Bids must be multiples of 10.
\item \textbf{Buy-right decision}: \texttt{\{sell, buy\_right\}}:as auctioneer, accept the highest
    bid (sell) or exercise the buy-right by paying the bid amount to the bidder.
\item \textbf{TC initiation}: \texttt{\{target, animal, offer\_cards\}}:select opponent and
    animal type, then select specific money cards from hand as the face-down offer (including 0-value
    cards for bluffing).
\item \textbf{TC defense}: \texttt{\{accept, counter(cards)\}}:accept the face-down offer
    (take money sight-unseen, surrender animals) or counter with specific money cards from hand.
\item \textbf{TC tie retry}: \texttt{\{offer\_cards\}}:after a tie, submit a new offer using
    specific cards. After three ties, the initiator wins by default.
\item \textbf{Scratchpad update}: \texttt{\{notes\}}:free-text strategic notes (full memory mode
    only).
\end{enumerate}

\paragraph{Error handling by action type.}
The framework validates all actions and applies action-specific recovery procedures.
Table~\ref{tab:errors} summarizes the handling for each error case.

\begin{table}[h]
\centering
\small
\begin{tabular}{p{3.2cm}p{3.5cm}p{5.5cm}}
\toprule
\textbf{Action Type} & \textbf{Error Case} & \textbf{Handling} \\
\midrule
All actions & Unparseable JSON & Multi-stage fallback: (1)~direct parse, (2)~regex extraction of last JSON object from free text, (3)~manual key-value extraction with type coercion \\
\addlinespace
All actions & Response truncated by token limit & Lightweight retry call requesting only the JSON action (512 tokens, reasoning disabled) \\
\addlinespace
All actions & API failure & Exponential backoff retry (1s, 2s, 4s) up to configurable max attempts \\
\midrule
Auction bid & Non-multiple of 10 & Rounded down to nearest multiple of 10 \\
\addlinespace
Auction bid & Bid below minimum & Clamped up to minimum bid increment \\
\addlinespace
Auction bid (canonical) & Bid exceeds wealth & \emph{Not corrected.} First overbid triggers wealth revelation and auction restart; a second overbid in the same auction eliminates the player \\
\addlinespace
Auction bid (fast) & Bid exceeds wealth & Clamped to total money \\
\midrule
Buy-right & Cannot afford payment & Action rejected; card sold to bidder \\
\midrule
TC initiation & Invalid target (wrong animal, self-trade) & Observation includes pre-computed valid targets; if the LLM names an invalid target or animal, the agent falls back to the first valid (target,~animal) pair \\
\addlinespace
TC initiation & Offer cards not in hand & One retry with error feedback showing available cards; fallback to greedy card selection \\
\midrule
TC defense & Counter cards not in hand & Same retry-then-greedy procedure as initiation \\
\midrule
Payment (all types) & Exact amount not possible & DP algorithm finds minimum-overpay card combination. No change given, matching tabletop rules \\
\bottomrule
\end{tabular}
\caption{Error handling by action type. The framework distinguishes between errors that are silently corrected (bid rounding), errors that trigger retries (malformed JSON, invalid cards), and errors that are game mechanics (overbidding). The canonical auction mode intentionally permits overbids because the resulting wealth revelation is a strategically meaningful penalty.}
\label{tab:errors}
\end{table}

\paragraph{Payment resolution.}
Because no change is given, the engine must select a subset of money cards whose sum meets or
exceeds the required amount while minimizing overpayment. This is a variant of subset sum solved via
dynamic programming over the player's money cards. The algorithm finds the smallest achievable sum
$\geq$ target and, among ties, the combination using the fewest cards. For example, paying 60 with
cards \{100, 50, 10, 10, 0, 0\} selects 50+10 (exact) rather than 100 (overpay by 40).

\paragraph{Scratchpad mechanism.}
In \emph{full} memory mode, each agent maintains a personal scratchpad~$s^i$, a free-form text
buffer capped at ${\sim}$300 tokens. At the end of a player's own turn, if new observable events
occurred, the framework issues an additional LLM call providing the current scratchpad contents
and the new events, and requests updated notes. The agent uses the \emph{same model} as for game
decisions, preserving reasoning style consistency. The updated scratchpad is included in all
subsequent observations under a ``\textsc{Your Notes}'' heading.

% Appendix D — Full agent prompt templates (from codebase)
\section{Agent Prompts}
\label{app:prompts}

All prompts below are reproduced verbatim from the benchmark implementation. Template variables (in
angle brackets) are filled dynamically from game state. Every action prompt ends with a standardized
JSON response instruction.

\subsection{System Prompt (Game Rules)}

Every LLM call uses a system prompt formed by concatenating the neutral game-rules block below
with the \texttt{optimal} character prompt in Appendix~\ref{app:prompts}. No strategy hints,
examples, or tactical advice are appended for our evaluations; the codebase supports
biased-character variants (e.g.\ \texttt{multiplier\_hunter}, \texttt{quartet\_closer}) for
ablation purposes only.

{\small
\begin{verbatim}
KUHHANDEL MASTER - RULES

GOAL: Complete quartets (4 cards of same animal).
Score = (sum of complete quartet values) × (number of quartets).
Example: 1 donkey quartet (500) + 1 horse quartet (1000)
         = (500+1000) × 2 quartets = 3000 points.
Example: 1 horse quartet alone = 1000 × 1 = 1000 points.
         Incomplete sets score 0.

ANIMALS (Complete Quartet Value):
Chicken=10, Goose=40, Cat=90, Dog=160, Sheep=250,
Goat=350, Donkey=500, Pig=650, Cow=800, Horse=1000

YOUR TURN - Choose one:

[A] AUCTION: Draw a card from the deck. Other players bid
    (each bid must exceed the previous by at least 10 coins).
    Then you (the auctioneer) choose:
    - ACCEPT highest bid: they pay you, get animal
    - BUY-RIGHT: you pay that amount TO highest bidder,
      YOU keep animal
    No change given - must pay exact or overpay.
    OVERBID: If the highest bidder cannot pay, they must
    reveal all their money cards. The auction restarts and
    the overbidder may not overbid again.

[B] KUHHANDEL (Cattle Trade): Trade an animal type you
    BOTH own with an opponent.
    - You place a hidden offer face-down (using coin cards;
      zeros allowed for bluffing)
    - Opponent chooses:
      - ACCEPT: takes your hidden coin cards (without seeing
        them first), gives you their animal
      - COUNTER: places their own face-down offer
    - If countered: heaps are swapped and counted secretly
      by each player. Whoever OFFERED more coins wins and
      takes the animal(s). The exact amounts and card counts
      stay private; other players learn only who won.
    - Ties: repeat with new offers up to 3x, then challenger
      wins automatically.
    - If both have 2+: ALL cards of that type trade.
\end{verbatim}
}

\subsection{Character Prompt}

The \texttt{optimal} character appended to the rules above is deliberately minimal: two lines
that name the goal without prescribing how to achieve it. This is the only character used in our
evaluations; together with the rules block above it constitutes the complete system prompt every
LLM receives:

{\small
\begin{verbatim}
You are an AI playing Kuhhandel Master. Your goal is to win.
Play optimally to maximize your expected score.
\end{verbatim}
}

\subsection{Observation Format}

At each decision point, the agent receives a structured observation:

{\small
\begin{verbatim}
=== GAME STATE (You are Player <id>) ===
Turn: <turn_number>
Cards remaining in deck: <deck_remaining>

YOUR HAND:
  Money: <total> coins total
  Money cards: 100 coins, 50 coins, 10 coins, 10 coins,
               0 coins, 0 coins
  Animals: 3x Horse, 2x Cow, 1x Sheep

OTHER PLAYERS:
  Player 1: 2x Dog, 1x Sheep | 5 coin cards
  Player 2: 1x Horse, 2x Chicken | 3 coin cards
  Player 3: 2x Cow, 1x Goat | 7 coin cards

VALID KUHHANDEL TARGETS:
  Player 2: horse (you have 3, they have 1)
  Player 3: cow (you have 2, they have 2)

RECENT EVENTS (what you observed):
  Turn 21: Player 1 won auction for dog (40 coins)
  Turn 22: Kuhhandel: Player 2 challenged Player 3
           for cow. Player 3 countered. Player 2 won.

YOUR NOTES:
  <scratchpad contents, full memory mode only>
\end{verbatim}
}

Opponent money card \emph{counts} are visible but not values, matching the tabletop game's
information structure. A player's exact wealth is revealed only through overbid events.

\subsection{Turn Choice}

When it is the active player's turn, they choose between auction and Kuhhandel:

{\small
\begin{verbatim}
<observation>

It's your turn. You must choose ONE action:
1. AUCTION - Flip a card from the deck and auction it
2. KUHHANDEL - Challenge another player to trade an
   animal you both own

Respond with JSON:
{"reasoning": "...", "action": "auction" or "kuhhandel"}
\end{verbatim}
}

\subsection{Canonical Auction Bidding}

In canonical mode, all non-auctioneer players submit bids simultaneously each round:

{\small
\begin{verbatim}
<observation>

CANONICAL AUCTION - Simultaneous Bidding Round
Round <n>
- Card: sheep (quartet value: 250)
- Current price: 30
- Current winner: Player 3
- You have: 1 of this animal
- Your money: 170 coins (100, 50, 10, 10, 0, 0)

All bidders submit simultaneously. To win, bid HIGHER
than current price. You may bid more than you can pay.
If you win and cannot pay, your total wealth is revealed
to all players and the auction restarts from 0. A second
overbid eliminates you from this auction.

IMPORTANT: Bids must be multiples of 10 coins.

Respond with JSON:
{"reasoning": "...", "action": "pass" or "bid",
 "amount": <multiple of 10>}
\end{verbatim}
}

The prompt explicitly states overbid consequences. Models that overbid reveal their wealth to all
opponents and are excluded from the restarted auction.

\subsection{Auctioneer Buy-Right Decision}

After bidding concludes, the auctioneer decides whether to sell or exercise buy-right:

{\small
\begin{verbatim}
<observation>

BUY-RIGHT DECISION (you are auctioneer)
- Card: sheep (quartet value: 250)
- Highest bid: 60 by Player 1
- You have: 1 of this animal
- Your money: 170 coins (100, 50, 10, 10, 0, 0)

Options:
- SELL: Take 60 coins, Player 1 gets the card
- BUY_RIGHT: Pay 60 coins to Player 1, YOU keep the card

Respond with JSON:
{"reasoning": "...", "decision": "sell" or "buy_right"}
\end{verbatim}
}

\subsection{Kuhhandel Initiation}

The active player selects a target and constructs a face-down offer from specific money cards:

{\small
\begin{verbatim}
<observation>

KUHHANDEL - Choose target and make offer:
Valid targets:
- Player 1: horse (you have 2, they have 1)
- Player 3: cow (you have 1, they have 2)

Remember:
- You're trading animals you BOTH own
- Your hidden offer is a set of money cards from your hand
- The opponent sees how many cards you offer, but not
  their values
- You can include 0-coin cards to make your offer look
  bigger (bluff)
- Opponent will either ACCEPT (take money, give animal)
  or COUNTER

Your available card values: [100, 50, 10, 10, 0, 0]
Select exactly which cards to include in your offer.

Respond with JSON:
{"reasoning": "...", "target_player": <id>,
 "animal": "<type>",
 "offer_cards": [10, 10, 0, 0]}
\end{verbatim}
}

The model selects concrete card values from its hand. The framework validates that the requested
cards exist; if not, it retries with an error message showing available cards.

\subsection{Kuhhandel Defense}

The target of a Kuhhandel chooses to accept the hidden offer or counter:

{\small
\begin{verbatim}
<observation>

KUHHANDEL - You are being challenged!
- Player 1 wants your cow
- You have 2 of this animal (quartet value: 800)
- Trade size: 2 card(s)
- Their offer is face-down (3 card(s), unknown value)

Options:
1. ACCEPT - Take their hidden money, give them your
   animal(s)
2. COUNTER - Select cards from your hand as counter-offer.
   Offers are revealed and exchanged. Winner = whoever
   OFFERED more money. Loser gives up animal(s).

Your available card values: [50, 10, 10, 0, 0]
If countering, select exactly which cards to include.

Respond with JSON:
{"reasoning": "...", "action": "accept" or "counter",
 "counter_cards": [50, 10, 10]}
\end{verbatim}
}

\subsection{Kuhhandel Tie Retry}

After a tie (equal offer values), both players make new offers:

{\small
\begin{verbatim}
<observation>

KUHHANDEL TIE - Make a new offer!
- Trading: cow with Player 3
- Previous offers tied (your offer was ~60)
- Tie count: 1 (after 3 ties, you win automatically)

Your available card values: [100, 50, 10, 10, 0, 0]
Select exactly which cards to include in your new offer.

Respond with JSON:
{"reasoning": "...",
 "offer_cards": [50, 10, 10, 0]}
\end{verbatim}
}

\subsection{Scratchpad Update}

In full memory mode, the framework issues an additional call after observable events to update the
agent's scratchpad:

{\small
\begin{verbatim}
Update your game notes based on these new events.
Keep under 300 tokens.
Focus on: opponent wealth signals, behavioral patterns,
quartet progress, trade outcomes.

Current notes:
<previous scratchpad or "(none yet)">

New events:
Turn 23: Player 1 won auction for dog (40 coins)
Turn 24: You initiated Kuhhandel with Player 2 for horse.
  You offered 50, they countered 0. You won, got 1 horse.
Turn 25: Player 3 completed quartet (sheep)

Respond with ONLY the updated notes, nothing else.
\end{verbatim}
}

The scratchpad is capped at approximately 300 tokens and uses the same model as game decisions. It
is included in all subsequent observations under the ``YOUR NOTES'' heading.

\end{document}